\documentclass[journal]{IEEEtran}
\usepackage{cite}

\ifCLASSINFOpdf
  \usepackage[pdftex]{graphicx}
\else
\fi
\usepackage{amsmath}
\usepackage{array}
\usepackage{amsthm}
\usepackage{amssymb}
\usepackage{color}
\usepackage{comment}
\usepackage[linesnumbered,ruled]{algorithm2e}
\usepackage{threeparttable}
\usepackage{multirow}
\usepackage{booktabs}
\usepackage{subfig}
\usepackage[shortlabels]{enumitem}

\usepackage{bm}
\usepackage{graphicx}

\usepackage{url}

\begin{document}
%
\title{Learning to Discover Knowledge: \\A Weakly-Supervised Partial Domain Adaptation Approach}
%
%
%

\author{~Mengcheng~Lan,
	Min~Meng,~\IEEEmembership{Member,~IEEE,}
	~Jun~Yu,~\IEEEmembership{Member,~IEEE,} \protect\\
	and~Jigang~Wu,~\IEEEmembership{Member,~IEEE} 
	\thanks{
		This work was supported in part by the National Natural Science Foundation of China 
		under Grant 62172109, Grant 62072118 and Grant 62020106007,
		in part by the Natural Science Foundation of Guangdong Province under Grant 2022A1515010322, 
		and in part by the Guangdong Basic and Applied Basic Research Foundation under Grant 2021B1515120010. 
		\emph{(Corresponding author: Min Meng.)}}
	\IEEEcompsocitemizethanks{\IEEEcompsocthanksitem M. Lan, M. Meng and J. Wu are with the Department
		of Computer Science, Guangdong University of Technology, Guangzhou 510006, China 
		(e-mail: lanmengchengds@gmail.com; mengmin1985@gmail.com; asjgwucn@outlook.com).
		\IEEEcompsocthanksitem J. Yu is with the School of Computer Science,
		Hangzhou Dianzi University, Hangzhou 310018, China (e-mail: zju.yujun@gmail.com).}
}

%
%

\markboth{IEEE Transactions on ,~Vol.~, No.~, ~2021}%
{Meng \MakeLowercase{\textit{et al.}}:}
%



\maketitle

\begin{abstract}
Domain adaptation has shown appealing performance by leveraging knowledge from a source domain with rich annotations. 
However, for a specific target task, it is cumbersome to collect related and high-quality source domains. 
In real-world scenarios, large-scale datasets corrupted with noisy labels are easy to collect, 
stimulating a great demand for automatic recognition in a generalized setting, i.e., weakly-supervised partial domain adaptation (WS-PDA), 
which transfers a classifier from a large source domain with noises in labels to a small unlabeled target domain. 
As such, 
the key issues of WS-PDA are: 
1) how to sufficiently discover the knowledge from the noisy labeled source domain and the unlabeled target domain,
and 2) how to successfully adapt the knowledge across domains. 
In this paper,
we propose a simple yet effective domain adaptation approach, termed as self-paced transfer classifier learning (SP-TCL), to address the above issues, 
which could be regarded as a well-performing baseline for several generalized domain adaptation tasks. 
The proposed model is established upon the self-paced learning scheme, seeking a preferable classifier for the target domain. 
Specifically, 
SP-TCL learns to discover faithful knowledge via a carefully designed prudent loss function 
and simultaneously adapts the learned knowledge to the target domain by iteratively excluding source examples from training under the self-paced fashion. 
Extensive evaluations on several benchmark datasets demonstrate that 
SP-TCL significantly outperforms state-of-the-art approaches on several generalized domain adaptation tasks. 
\end{abstract}


%
\IEEEpeerreviewmaketitle

\section{Introduction}\label{sec:introduction}
Due to vast quantities of data, 
deep neural networks have gained impressive performance for a variety of machine perception tasks. 
In spite of their success, 
they usually require massive amounts of annotated training samples, which is often prohibitive in real-world scenarios. 
Thus, it is natural to leverage rich annotations from related datasets (source domain) 
to facilitate the task of interest (target domain) where the label is absent or cost-expensive. 
However, 
such an intuitive strategy will degenerate or even fail due to the distribution shift from the source versus target domains. 
Domain adaptation (DA) \cite{pan2009survey}, which attempts to reduce the domain divergence, 
has provided appealing solutions for numerous applications, 
such as image classification \cite{Long2014Transfer}, person re-identification \cite{deng2018image}, semantic segmentation \cite{lan2023smooseg}. 

\begin{figure}[t]
	\centering
	\includegraphics[width=3.3in]{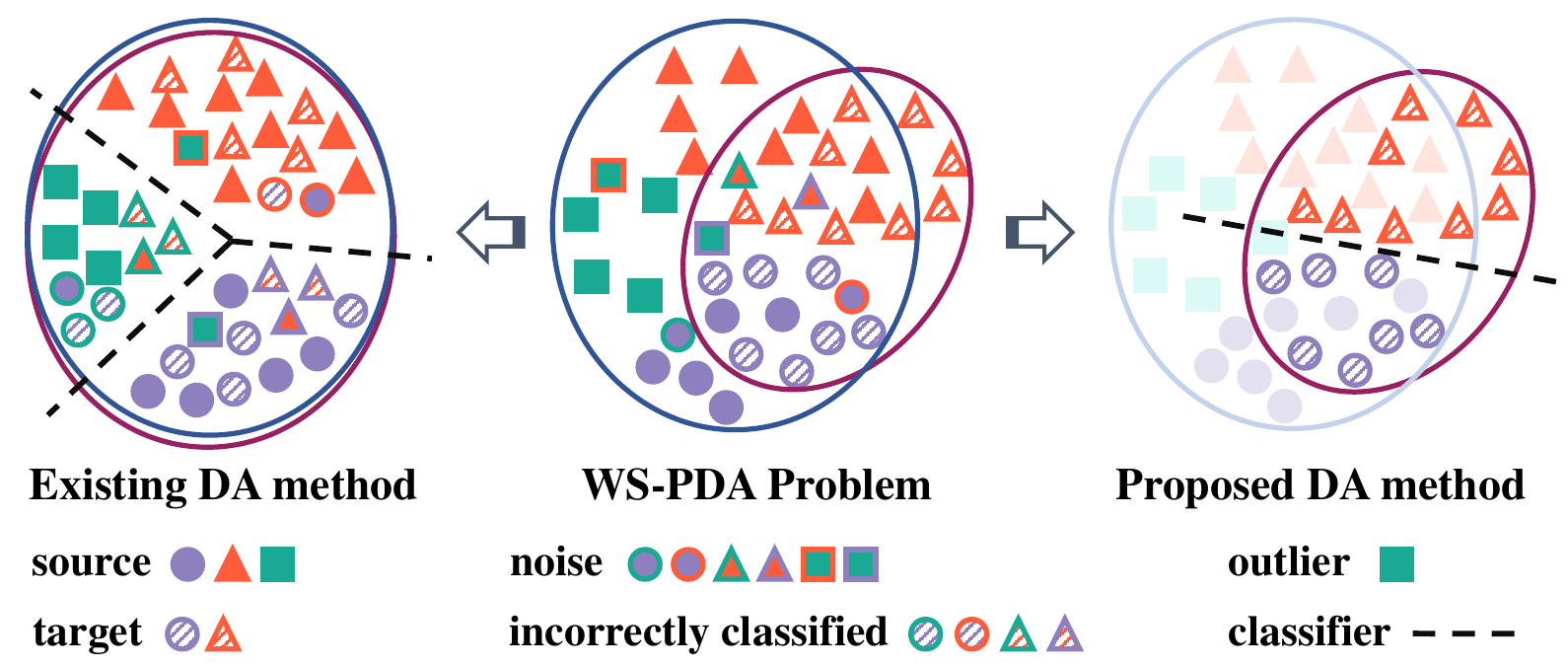}
	\captionsetup{font={footnotesize}}
	\caption{The problem of weakly-supervised partial domain adaptation, 
		where the target label space is a subspace of the source label space and the source domain is naturally corrupted with noisy labels. 
		For each noisy or incorrectly classified sample, the face-color denotes its ground truth, while the edge-color denotes its noisy label. 
		For existing traditional DA methods, outlier classes will cause negative transfer, and noisy samples will affect the final classifier learning. 
		In our SP-TCL, we propose knowledge discovery and knowledge adaptation strategies to effectively address the problem of weakly-supervised partial domain adaptation.
	}
	\label{fig:WS-PDA problem}
\end{figure}

Existing studies on domain adaptation generally assume that 
the source and target domains share identical label space, and all source examples are labeled with accurate annotations. 
Considerable research efforts have been made to establish knowledge transfer from well-labeled source domain to a non-labeled target domain, 
including learning transferable representation via feature transformation \cite{pan2010domain,Long2014Transfer,Tian2020Domain}, 
learning transferable instances via estimating their importances \cite{sugiyama2008direct}, 
and learning an adaptive classifier based on the structural risk minimization principle \cite{Duan2012Domain,Lorenzo2010Domain,wang2018visual}. 

However, in real-world applications, 
it is easy to encounter situations where the aforementioned assumptions fail.
In practical scenarios, 
it is difficult to collect a source dataset with its label space identical to the target dataset of interest, 
since we have no prior knowledge of the label space on the target domain. 
A more feasible way is 
to find a large-scale source dataset, e.g., ImageNet-1K \cite{russakovsky2015imagenet}, that is diverse enough to contain all categories of target domain, 
which leads to partial domain adaptation (PDA) \cite{cao2018partial_PADA,cao2019learning, chen2020domain1} that performs knowledge transfer from a many-class domain to a few-class domain. 
Besides, manually collecting large-scale labeled datasets is usually cost-expensive and labor-intensive. 
A widely used surrogate is to automatically collect the noisy labeled datasets from Internet or social media,
which results in weakly-supervised domain adaptation (WS-DA) \cite{shu2019transferable}. 
Such two scenarios are more general and challenging than standard domain adaptation, 
and more likely to occur simultaneously in real-world applications. 

With the rapid development of learning tasks in new domains, 
we consider a novel and more applicable scenario of domain adaptation 
where the target label space is a subspace of the source label space, i.e., $\mathcal{Y}_t\subset\mathcal{Y}_s$,
and the source domain is naturally corrupted with noisy labels. 
Such a challenging scenario is referred to as \emph{weakly-supervised partial domain adaptation (WS-PDA)}, as shown in Fig. \ref{fig:WS-PDA problem}. 
In the new scenario, 
existing domain adaptation methods may suffer from two critical limitations: 
1) minimizing the distribution divergence across domains will lead to severe negative transfer due to the difference in label spaces; 
2) noisy examples will seriously affect the generalization ability of the classifier trained solely on the noisy source domain. 
It is also worth noting that, training the classifier to work well on the source domain will hurts the target performance 
if the existence of an optimal classifier with low generalization error on both source and target domains is not guaranteed \cite{Rui2018A}. 

In this paper, 
we present a novel weakly-supervised partial domain adaptation approach to address the above challenges. 
Inspired by \cite{Kumar2010Self,Tang2012Shifting}, 
we integrate transfer classifier learning and self-paced learning into a unified framework for knowledge discovery and adaptation in a generalized setting. 
Specifically, 
we introduce a prudent loss function to automatically discover faithful knowledge from both the noisy labeled source domain and the pseudo labeled target domain,  
which facilitates learning a joint classifier for source and target domains. 
Meanwhile, 
we adopt a self-paced regularizer for knowledge adaptation by gradually excluding complex (noisy or outlier) source examples from training, 
which realizes the transition from a joint classifier to a preferable classifier of target domain. 
Such a natural transition process allows us to progressively refine the classifier's decision boundaries to better fit the target distribution. 
With a joint learning manner, 
such two learning procedures benefit each other to improve the process of knowledge discovery and adaptation. 
The proposed method is applicable to several different domain adaptation scenarios, which relaxes the assumptions of most existing approaches. 
Experimental results on three benchmark datasets demonstrate the superiority of our approach on several generalized domain adaptation tasks. 
The main contributions of this paper are summarized as follows:
\begin{itemize}
	\item 
	To the best of our knowledge, 
	we are among the first to consider the weakly-supervised partial domain adaptation problem, 
	which transfers a classifier from a large source with noisy labels to a small unlabeled target domain. 
	We take DA a step further toward unsupervised learning in practical scenarios.
	
	\item
	We propose a simple yet effective approach to weakly-supervised partial domain adaptation, 
	which could be regarded as a well-performing baseline for several generalized domain adaptation tasks. 
	Specifically, our model aims to discover faithful knowledge from source and target domains via a prudent loss function 
	and simultaneously adapt the learned knowledge to the target domain by gradually excluding sources examples from training under the self-paced learning fashion. 
	
	\item 
	Extensive experimental results on several standard benchmarks demonstrate that
	the proposed model outperforms other state-of-the-art methods with remarkable margins in several generalized adaptation scenarios. 
	
\end{itemize}

The remainder of this paper is organized as follows. 
Section \ref{sec:relatedwork} briefly reviews some related literatures and highlights the differences of our method. 
In Section \ref{sec:method}, we introduce the proposed model and its optimization algorithm. 
Experimental results for several generalized adaptation tasks are presented in Section \ref{sec:experiments}. 
Finally, we conclude this paper in Section \ref{sec:conclusion}.

\section{Related Work}\label{sec:relatedwork}
\subsubsection{Traditional Domain Adaptation}
Over the past decade,
a large group of domain adaptation approaches \cite{pan2010domain,Long2014Transfer,Duan2012Domain,wang2018visual} have been proposed to overcome the generalization bottleneck caused by the domain shift. 
To mitigate the distribution shift between different domains, 
previous studies for unsupervised domain adaptation mainly focus on the following two aspects.
One is to learn transferable representation by feature transformation \cite{pan2010domain,luo2020unsupervised,Long2014Transfer,zhang2017joint,liang2018aggregating, meng2020coupled, Chen2020Domain,Li2020Locality}.
These methods aim to discover a common domain-invariant feature
space via minimizing the divergence between different domains under the Maximum Mean Discrepancy (MMD) \cite{gretton2007kernel} criterion.
As a typical work,
Joint Distribution Adaptation (JDA) \cite{Long2014Transfer} adopts MMD
for joint distribution alignment such that both the marginal and conditional distributions can be effectively matched in the common feature space.
\cite{liang2018aggregating} extends JDA by introducing the domain-irrelevant class clustering term to promote the intra-class compactness of data. 
The other one is to learn a transfer classifier based on the structural risk minimization principle \cite{Duan2012Domain,Lorenzo2010Domain,Long2014Adaptation,wang2018visual, meng2022dual}. 
Long \emph{et al.} \cite{Long2014Adaptation} propose a novel adaptive classifier learning approach by incorporating SRM and regularization theory into a general framework,
which significantly boosts the classification performance.

In light of the success of deep learning techniques, 
deep networks have been adopted to generate transferable representation for domain adaptation \cite{tzeng2014deep,long2015learning,long2017deep,Li2020Generating}, 
yielding evident performance improvement against shallow approaches.
Tzeng \emph{et al.} \cite{tzeng2014deep} introduce an additional domain confusion loss by utilizing MMD on the adaptation layer.
Furthermore, Joint Adaptation Networks (JAN) \cite{long2017deep} proposes a joint maximum mean discrepancy (JMMD) criterion to further minimize the domain shift by directly matching the joint distribution.
In addition,
adversarial learning-based methods \cite{goodfellow2014generative,tzeng2017adversarial,long2018conditional,tang2020discriminative} achieve considerable advances owing to their high-capacity of feature extraction functions.
Tzeng \emph{et al.} \cite{tzeng2017adversarial} present a novel generalized framework for adversarial adaptation learning.
Long \emph{et al.} \cite{long2018conditional} develop an adversarial domain adaptation framework to condition the adversarial domain adaptation on discriminative information.

Despite their promising results on UDA tasks,
the above approaches would inevitably fail when dealing with more challenging weakly-supervised partial domain adaptation,
in which the noisy source samples or samples belong to outlier classes will easily trigger the notorious negative transfer.

\subsubsection{Generalized Domain Adaptation}
Existing studies on domain adaptation generally assume that the label spaces of source and target domains are identical, 
and all source examples are labeled with accurate annotations, 
which may be restrictive for many real-world applications. 
To relax the assumption of identical label space, 
some partial domain adaptation (PDA) approaches \cite{cao2018partial_PADA,cao2019learning,li2020dual,li2020deep,ren2020learning, liang2021domain},
which perform knowledge transfer from the many-class domain to the few-class domain, 
have recently been proposed. 
A common practice of PDA is to develop a weighting mechanism for automatically distinguishing between shared classes and outlier classes, 
then simultaneously promote positive transfer of shared class examples as well as alleviate negative transfer of outlier class examples. 
For example,
The works in \cite{cao2018partial_PADA,li2020dual,li2020deep} propose to estimate the weight of each source class by averaging the predictions on all target data.
Cao \emph{et al.} \cite{cao2018partial_PADA} incorporate the learned class weight vector into the source classifier learning and the partial source adversarial domain discriminator learning.
Li \emph{et al.} \cite{li2020deep} utilize the class weight to perform weighted class-wise distribution matching.
Ren \emph{et al.} \cite{Ren2018Learning} adopt the class weight for partial feature alignment and source-specific classifier learning.
Besides,
The works in \cite{zhang2018importance,cao2019learning} suggest to learn the weight of each source example that quantifies the example’s transferability rather than each source class. 
To relax the assumption of clean source data, 
Shu \emph{et al.} \cite{shu2019transferable} develop a transferable curriculum learning (TCL) approach for weakly-supervised domain adaptation,
where the source data may contain label noise and/or feature noise.
TCL prioritizes the noiseless and transferable source examples to enhance positive transfer and simultaneously mitigate the negative transfer of noisy source examples. 
Liu \emph{et al.} \cite{liu2019butterfly} present a robust one-step approach to address the wildly unsupervised domain adaptation.

These pioneering approaches achieve significant performance improvement for partial or weakly-supervised domain adaptation tasks, respectively.
However, we believe that these two challenging generalized domain adaptation problems are more likely to be entangled in practical applications,
leading to the performance degradation of existing methods.

\subsubsection{Self-Paced Learning}
Inspired by the learning process of humans/animals,
Kumar \emph{et al.} \cite{Kumar2010Self} propose a novel learning paradigm, 
self-paced learning (SPL), 
which usually contains a weighted loss term on all examples and a SPL regularizer imposed on the weights of examples. 
The core idea of SPL is to learn the model incrementally using examples from easy to complex that are dynamically determined by the feedback of the learner itself.
By virtue of its generality,
various applications based on this theory have recently been proposed,
such as matrix factorization \cite{Qian2015Self},
multiple instance learning \cite{Zhang2015A},
multi-view clustering \cite{Chang2015Multi},
and domain adaptation \cite{Tang2012Shifting,shu2019transferable}.
The work in \cite{Tang2012Shifting} introduces a domain adaptation approach that iteratively includes examples from the target domain
and simultaneously excludes examples from the source domain to train the model in a self-paced fashion.
TCL \cite{shu2019transferable} proposes a novel latent weighting scheme to select easy and transferable source examples for model training.
Different from these works \cite{Tang2012Shifting,shu2019transferable}, 
we focus on the weakly-supervised partial domain adaptation scenario, 
where the domain shift exists along with noisy/outlier examples in the source domain.

\section{Proposed Method}\label{sec:method}
\subsection{Problem Definition}
In this paper, 
we consider a novel and more applicable domain adaptation scenario, \emph{weakly-supervised partial domain adaptation}. 
We assume that there is a source domain $\mathcal{D}_s=\{(x_i^s,y_i^s)\}_{i=1}^{n_s}$ with $n_s$ labeled examples associated with $|\mathcal{Y}_s|$ classes 
and a target domain $\mathcal{D}_t=\{x_i^t\}_{i=1}^{n_t}$ with $n_t$ unlabeled examples associated with $|\mathcal{Y}_t|$ classes. 
$y_i^s\in\mathcal{R}^{|\mathcal{Y}_s|}$ is the one-hot encoded label vector of the source example $x_i^s\in\mathcal{R}^{m}$. 
For simplicity, 
we denote $X_s=[x_1^s,\cdots,x_{n_s}^s]\in\mathcal{R}^{m\times n_s}$ and $X_t=[x_1^t,\cdots,x_{n_t}^t]\in\mathcal{R}^{m\times n_t}$ 
as the source and target data matrices, respectively. 
Particularly, in weakly-supervised partial domain adaptation, 
we relax the assumption of standard domain adaptation to that the source domain may be partially corrupted with noisy labels 
and the target label space is a subspace of the source label space, i.e., $\mathcal{Y}_t\subset\mathcal{Y}_s$. 
Let $\mathcal{P}$ and $\mathcal{Q}$ denote the marginal distributions of source and target domains, respectively,
we further have $\mathcal{P}_{\mathcal{Y}_t}\not = \mathcal{Q}$ in the present of domain shift,
where $\mathcal{P}_{\mathcal{Y}_t}$ represents the distribution of source data that belongs to the label space $\mathcal{Y}_t$.
The goal of this work is to learn a transfer classifier $f:X_t\to Y_t$ using both the noisy labeled source data and unlabeled target data 
to predict the labels $\widetilde{Y}_t\in\mathcal{Y}_t$ for target domain $\mathcal{D}_t$. 

There are two critical technical difficulties for solving the weakly-supervised partial domain adaptation problem. 
On one hand, 
without access to the target labels, 
it is cumbersome to select the source examples from the shared label space $\mathcal{Y}_s\bigcap\mathcal{Y}_t$ between source and target domains, 
which indicates that performing distribution adaptation is problematic and will not benefit the target tasks. 
On the other hand, 
the source examples that are corrupted with noisy labels or belong to the outlier label space $\mathcal{Y}_s / \mathcal{Y}_t$ will cause severe \emph{negative transfer}. 
This problem is more general and challenging than partial domain adaptation, 
since the classification models are prone to overfitting these irrelevant and noisy source examples, 
resulting in their generalization performance degradation on the target domain. 
As such, 
the two challenges in distribution shift and noisy/outlier examples are entangled, 
making existing domain adaptation approaches infeasible. 

\subsection{Problem Formulation}
 
The crucial issues of WS-PDA can be abstracted as follows:  
1) how to sufficiently discover the knowledge from noisy labeled source domain and unlabeled target domain,
and 2) how to successfully adapt the knowledge across domains, 
on account of which the classification model can be expected to generalize well on target domain. 
In this paper, 
we introduce a novel self-paced transfer classifier learning (SP-TCL) framework to address the above issues, 
which is designed from two strategies: 1) \emph{generalized knowledge discovery} and 2) \emph{generalized knowledge adaptation}. 
As a result, our learning model can be formulated as: 
\begin{equation}
\begin{split}
\mathop{\arg\min}_{f\in\mathcal{H}_K,v,P_s,P_t,}\sum_{i=1}^{n_s}v_i\mathcal{L}_s(x^s_i,p_i^s;f)+\sum_{j=1}^{n_t}\mathcal{L}_t(x^t_j,p_j^t;f)\\
+g(v,\lambda)+\eta\left\| f\right\|^2_K+\rho\left\| f\right\|^2_M
\end{split}
\label{eq:learning model}
\end{equation}
where $f$ is the classifier, 
$\left\|f\right\|_K^2$ and $\left\|f\right\|_M^2$ are regularization terms that are used to control the complexity and smoothness of the classifier respectively. 
$\eta$ and $\rho$ are positive regularization parameters. 
$\mathcal{L}_{s/t}$ denotes a certain loss function. 
$g(v,\lambda)$ is the self-paced regularizer and $v\in {[0,1]}^{n_s}$ is the weight vector on all source data.

\subsubsection{Generalized Knowledge Discovery} 

To discover faithful knowledge from noisy labeled source domain and pseudo-labeled target domain, it is critical to design
an appropriate loss function to distinguish between easy and
complex examples. 
An intuitive way is to adopt the square loss $\ell_2$ on all data as $\mathcal{L}_s(x^s_i,y_i^s;f)=(f(x_i^s)-y_i^s)^2$ and $\mathcal{L}_t(x^t_i,\widetilde{y}_i^t;f)=(f(x_i^t)-\widetilde{y}_i^t)^2$,
where $\widetilde{y}_i^t$ is the pseudo label (hard label) of target example $x^t_i$. 
However, such an 'aggressive' loss function is not suitable for the WS-PDA problem, 
since it cannot deal with the noisy labels on source domain and tends to overfit the source dataset. 
To cope with this issue, 
we introduce a 'prudent' loss function using the class probability matrix instead of hard labels as follows: 
\begin{equation}
\mathcal{L}_{s/t}(x^{s/t}_i,p_i^{s/t};f)=\sum_{c=1}^{|\mathcal{Y}_s|}[p_{ci}^{s/t}]^r(f(x_i^{s/t})-e_c)^2
\end{equation}
where $e_c=[0,\cdots,1,\cdots,0]\in\mathcal{R}^{|\mathcal{Y}_s|}$ is the one-hot encoded class indicator vector for the $c$-th class. 
$P_s\in\mathcal{R}^{|\mathcal{Y}_s|\times n_s}$ and $P_t\in\mathcal{R}^{|\mathcal{Y}_s|\times n_t}$ are the class probability matrices of the source domain and the target domain, respectively,
with each element $p_{ci}^{s/t}$ measuring the probability that the $i$-th source/target example belongs to the $c$-th class.
The class probability matrices $P_{s/t}$ should satisfy the constraints: $p_{ci}^{s/t}\ge 0, \sum_{c=1}^{|\mathcal{Y}_{s}|}p_{ci}^{s/t}=1, \forall i$. 
$r\geq1$ is the exponent of $p_{ci}^{s/t}$ that is used to adjust the probability distribution. 

We show that $P^{s/t}$ can automatically distinguish between the boundary points (complex examples) and clearly classified points (easy examples), 
and simultaneously adaptively weaken the effect of boundary points while maintaining the contribution of clearly classified points to classifier learning. 
Without losing generality, considering a 3-class classification
problem and $r=2$. 
For the clearly classified data points, the element values of $p_i$ will show significant difference. 
Assume $p_i=[0.8,0.1,0.1]^T$, then $p_i^r=p^2_i=[0.64,0.01,0.01]^T$,
in which case the data points still have large weights and contribute a lot to the classifier learning.
In contrast, the element values of $p_i$ tend to be equal for boundary data points which usually are noisy source data or falsely pseudo-labeled target data,
e.g., $p_i=[0.4,0.3,0.3]$, then $p_i^r=p^2_i=[0.16,0.09,0.09]^T$.
Thence, the weights of boundary points would be suppressed more than those of clearly classified points.

With such an iterative classifier training strategy, 
the falsely labeled examples are expected to progressively approach their true labels when updating the class probability matrices.
Therefore, due to the flexibility of the prudent loss function,
the proposed model can sufficiently extract faithful knowledge from noisy labeled source domain and pseudo-labeled target domain, 
which can reveal the discriminative structure knowledge to the transfer classifier learning and provide reliable guidance for target data annotation.
Please note that, by explicitly incorporating unlabeled target data into classifier learning, 
our method can iteratively refine the decision boundary to be more discriminative to the data distribution of both domains,
resulting in a joint classifier for data annotation.

\subsubsection{Generalized Knowledge Adaptation}
\begin{figure}[t]
	\centering
	\includegraphics[width=3.3in]{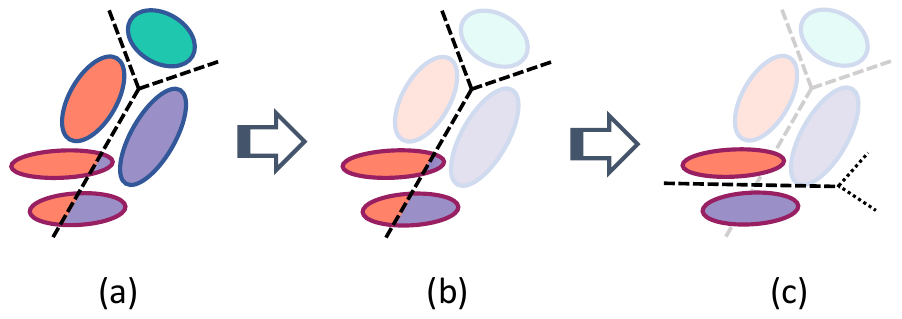}
	\captionsetup{font={footnotesize}}
	\caption{The overview of generalized knowledge adaptation: (a) the original joint classifier, (b) self-pace learning excludes source data from training, (c) manifold learning enforces the decision boundary to locate in the low-density area of target data, resulting in target-preferable classifier.}
	\label{fig: GKA}
\end{figure}
 
Building on the discovered faithful knowledge from source and target domains, 
we take a self-paced learning scheme \cite{Kumar2010Self} for weakly-supervised partial domain adaptation,
which adapts the classifier from large-scale noisy labeled source dataset to unlabeled target dataset. 
Specifically, SP-TCL aims to distinguish easy (i.e., confident and relevant) examples and then gradually transfer the knowledge to facilitate the target tasks, 
which realizes the transition from a joint classifier to a target-preferable classifier. 
The overview of generalized knowledge adaptation is illustrated in Fig. \ref{fig: GKA}.

In practice, 
source examples with large losses often locate around decision boundaries, 
which are usually confusing for the classifier learning. 
As such, we would like to stop learning from such noisy/outlier examples as they might cause negative effects on the transfer classifier.
Formulating this intuition, 
we introduce a self-paced regularizer to dynamically select the examples to learn from, 
which is designed as:
\begin{equation}
g(v,\lambda)=-\lambda\left\|v\right\|_1=-\lambda\sum_{i=1}^{n_s}v_i
\label{eq:self-paced regularizer}
\end{equation}

At the beginning, 
the age $\lambda$ of the model is large, the model prefers to consider using all source examples in classifier learning. 
With gradually decreasing $\lambda$, 
the model tends to select easy examples with smaller losses in favor of complex examples with large losses. 
By repeating this process, 
the model iteratively decreases examples from the source domain 
while leaving the classifier certain freedom to fit the target distribution, thereby establishing more confidence on the target domain. 
When all source examples are excluded from the classifier learning,
we complete the transfer classifier adaptation process and successfully adapt the classification model to the target domain. 

It is worthy noting that we do not integrate the self-paced learning regime into the target domain, 
which stems from two main reasons. 
Firstly, in the early phase, the classifier learned from a large amount of source examples will introduce certain domain bias, 
which is not suitable for distinguishing between easy and complex examples on the target domain. 
Secondly, the classifier is prone to overfitting a few examples on the target domain in the early phase, 
and such phenomenon will become more and more severe during the transfer process. 
Therefore, all target examples are included in the learning throughout the classifier adaptation process. 

Moreover, 
we proceed the classifier transfer by constantly excluding source examples from the learning process, 
which provides the classifier with maximum degrees of freedom to fit the target distribution.  
Therefore, we need to develop some criteria to promote the classifier adaptation process. 
Manifold learning has proven its effectiveness for semi-supervised learning in many applications. 
According to the smoothness assumption \cite{belkin2006manifold},
two close data points $x_1,x_2\in \mathcal{X}$ in the intrinsic geometry $\mathcal{Q}(\mathcal{X})$ tend to have similar conditional distributions.
In other words, 
the output label of prediction function varies smoothly along the geodesics in the inherent structure of data. 
Hence, we further exploit the knowledge of marginal distribution of target domain to facilitate transfer classifier learning, as shown in Fig. \ref{fig: GKA} (c).
Specifically, we aim at solving the following manifold learning problem: 
\begin{equation}
\left\|f\right\|_M^2=
\sum_{i,j=1}^{n_t}(f(x_i^t)-f(x_j^t)^2M_{ij}=\sum_{i,j=1}^{n_t}f(x_i^t)L^t_{ij}f(x_j^t)
\end{equation}
where $M$ is the affinity matrix, 
$L_t=D-M$ is the graph Laplacian matrix of $M$ and $D$ is a diagonal matrix with $i$-th diagonal element $d_{ii}=\sum_j M_{ij}$.
The affinity matrix $M$ can be calculated as:
\begin{equation}
M_{ij}=\begin{cases}
\text{cos}(x^t_i,x^t_j),\quad&\text{if}\ x^t_i\in\mathcal{N}_k(x^t_j)\ \text{or}\ x^t_j\in\mathcal{N}_k(x^t_i)\\
0,\quad&\text{otherwise}
\end{cases}
\label{eq:affinity matrix}
\end{equation}
where $\mathcal{N}_k(x^t_i)$ denotes the set of $k$-nearest neighbors of sample $x^t_i$, and we set $k=5$ in our experiments.
Inspired by \cite{belkin2006manifold},
we adopt the normalized Laplacian $L_t\leftarrow D^{-1/2}L_tD^{-1/2}$ in our formula.

\subsubsection{Overall Formulation}

We utilize a simple linear function $f(x_i)=W^Tx_i$ and reformulate our final model as follows: 
\begin{equation}
\begin{split}
\mathop{\arg\min}_{W,v\in[0,1]^{n_s},P_s,P_t,}&\sum_{i=1}^{n_s}v_i\sum_{c=1}^{|\mathcal{Y}_s|}[p_{ci}^s]^r\left\|W^Tx_i^s-e_c\right\|_2^2\\
+&\sum_{j=1}^{n_t}\sum_{c=1}^{|\mathcal{Y}_s|}[p_{cj}^t]^r\left\|W^Tx_j^t-e_c\right\|_2^2+\eta\left\| W\right\|^2_F\\
+&\rho Tr(W^TX_tL_tX_t^TW)-\lambda\sum_{i=1}^{n_s}v_i\\
s.t. \; &p_{ci}^s\ge0,\ \sum_{c=1}^{|\mathcal{Y}_s|}p_{ci}^s=1,\ \forall i=1,2,\cdots,n_s,\\
&p_{cj}^t\ge0,\ \sum_{c=1}^{|\mathcal{Y}_t|}p_{cj}^t=1,\ \forall j=1,2,\cdots,n_t
\end{split}
\label{eq:final model}
\end{equation}

\subsection{Optimization Strategy}

\begin{algorithm}[t]
	\caption{Optimization Algorithm for SP-TCL}
	\label{proposed_alg}
	\SetAlgoNoLine
	\SetKwInOut{Input}{\textbf{Input}} \SetKwInOut{Output}{\textbf{Output}}
	\Input{Source data $X_s$ with labels $Y_s$,
		target data $X_t$, 
		parameters $r$, $\rho$, $\eta$.}
	\Output{The classifier $f$ and the pseudo labels of target data $\widetilde{Y}_t$.}
	initialization: $P_s=Y_s$, $P_t=\bm{0}$;\\
	construct the kernel matrix $K=[k(x_i,x_j)]_{n\times n}$;\\
	construct the affinity matrix $M$ using \eqref{eq:affinity matrix};\\
	\While{not convergence}{
		\Repeat{convergence or reach the maximum iterations}{
			update $W$ using \eqref{eq:W-solve} or \eqref{eq:kernelization-solve};\\
			update $P$ using \eqref{eq:p-solve with r=1} or \eqref{eq:p-solve with r>1};\\}
		update latent weight variable $v$ using \eqref{eq:v-solve};\\
		update the learning pace $\lambda$;\\
	}
	$\widetilde{Y}_t=W^TX_t$ or $\widetilde{Y}_t=W^TK_t$.
\end{algorithm}

We employ alternative search strategy (ASS) to solve Eq. \eqref{eq:final model},
which optimizes one variable with the other variables fixed in an alternate manner. 
The overall optimization procedure of our model is summarized in Algorithm \ref{proposed_alg}. 

\emph{\textbf{Optimize} $\bm{v}$: }
For optimizing the latent weight variable $v$, we have
\begin{equation}
v^*=\mathop{\arg\min}_{v\in[0,1]^{n_s}}\sum_{i=1}^{n_s}v_il_i-\lambda\sum_{i=1}^{n_s}v_i
\label{eq:v-subproblem}
\end{equation}
where $l_i=\sum_{c=1}^{|\mathcal{Y}_s|}[p_{ci}^s]^r\left\|W^Tx_i^s-e_c\right\|_2^2$.
The global optimum $v^*=[v_1^*,\cdots,v_{n_s}^*]$ can be expressed as: 
\begin{equation}
v_i^*=\begin{cases}
1,\quad&l_i<\lambda\\
0,\quad&\text{otherwise}
\end{cases}
\label{eq:v-solve}
\end{equation}

\emph{\textbf{Optimize} $\{\bm{W,P_s,P_t}\}$: }
When fixing the latent weight vector $v$, 
we should jointly optimize other three variables $W$, $P_s$ and $P_t$ until convergence, 
where we can explore an EM-like optimization scheme. 
For \emph{E-step}, we fix $W$ and update $P_s$ and $P_t$, 
while for \emph{M-step}, we update the model parameter $W$ using updated $P_s$ and $P_t$. 

\textbf{E-step}: 
By removing the irrelevant terms w.r.t. $P_s$ and $P_t$,
we can reformulate Eq. \eqref{eq:final model} as: 
\begin{equation}
\begin{split}
\mathop{\arg\min}_{P}&\sum_{i=1}^{n}\sum_{c=1}^{|\mathcal{Y}_s|}u_ip_{ci}^r\left\|W^Tx_i-e_c\right\|_2^2\\
&s.t.\  p_{ci}\ge0,\ \sum_{c=1}^{|\mathcal{Y}_s|}p_{ci}=1,\ \forall i
\end{split}
\label{eq:P-problem}
\end{equation}
where 
$P=[P_s,P_t]\in\mathcal{R}^{|\mathcal{Y}_s|\times n}$,
$X=[X_s,X_t]\in\mathcal{R}^{m\times n}$,
$u=[v,\bm{1}_{1\times n_t}]$.
Let $q_{ci}=\left\|W^Tx_i-e_c\right\|_2^2$,
the problem \eqref{eq:P-problem} can be decomposed into $n$ independent subproblems as:
\begin{equation}
\begin{split}
\mathop{\arg\min}_{p_i}&\sum_{c=1}^{|\mathcal{Y}_s|}p_{ci}^rq_{ci},\quad s.t.\  p_{ci}\ge0,\sum_{c=1}^{|\mathcal{Y}_s|}p_{ci}=1,\ \forall i
\end{split}
\label{eq:P-subproblem}
\end{equation}
Note that the optimal solution of Eq. \eqref{eq:P-subproblem} is a specific solution of Eq. \eqref{eq:P-problem} as we set $u=\bm{1}_{1\times n}$.

When $r=1$, the optimal solution of Eq. \eqref{eq:P-subproblem} is given by: 
\begin{equation}
p_{ci}^*=\begin{cases}
1,\ &\text{if}\ c=\mathop{\arg\min}_{k}q_{ki}\\
0,\ &\text{otherwise}
\end{cases}
\label{eq:p-solve with r=1}
\end{equation}

When $r>1$, we have the Lagrangian function of Eq. \eqref{eq:P-subproblem}
\begin{equation}
\mathcal{L}_\theta=\sum_{c=1}^{|\mathcal{Y}_s|}p^r_{ci}q_{ci}-\theta(\sum_{c=1}^{|\mathcal{Y}_s|}p_{ci}-1)
\label{eq:p-Lagrangian function}
\end{equation}
where $\theta$ is the Lagrangian multiplier.
By setting the derivative of Eq. \eqref{eq:p-Lagrangian function} w.r.t. $p_{cj}$ to zero and combining with the constraint $\sum_{c=1}^{|\mathcal{Y}_s|}p_{ci}=1$,
we obtain the optimal solution of problem \eqref{eq:P-subproblem} as: 
\begin{equation}
p_{ci}^*={({\frac{1}{q_{ci}})}^{\frac{1}{r-1}}}/\sum_{c=1}^{|\mathcal{Y}_s|}{(\frac{1}{q_{ci}})}^{\frac{1}{r-1}}
\label{eq:p-solve with r>1}
\end{equation}
Then, we have $P_s=P_{1:n_s}$ and $P_t=P_{n_s+1:n}$.

\textbf{M-step}: 
Learning model parameter $W$ by fixing the other variables.
According to Eq. \eqref{eq:P-problem},
we rewrite the loss terms in Eq. \eqref{eq:final model} into a compact matrix representation: 
\begin{equation}
\begin{split}
&\sum_{i=1}^{n}\sum_{c=1}^{|\mathcal{Y}_s|}u_ip_{ci}^r\left\|W^Tx_i-e_c\right\|_2^2\\
=&\sum_{i=1}^{n}Tr(W^Tx_ix_i^TW)\sum_{c=1}^{|\mathcal{Y}_s|}u_ip_{ci}^r-2\sum_{i=1}^{n}u_i[p_i^r]^TW^Tx_i+\Omega\\
=&Tr(W^TXSX^TW)-2Tr(F^TW^TX)+\Omega
\end{split}
\label{eq:compact matrix}
\end{equation}
where $\Omega=\sum_{i=1}^{n}\sum_{c=1}^{|\mathcal{Y}_s|}u_ip_{ci}^r$ is a constant w.r.t. $W$.
$F=P^rU$, where $r$ applies element-wise exponential operations and $U=diag(u)$ is a diagonal matrix.
$S$ is a diagonal matrix with $s_{ii}=\sum_{c=1}^{|\mathcal{Y}_s|}F_{ci}$.
By substituting Eq. \eqref{eq:compact matrix} and adding a constant term $tr(S^{-1}F^TF)$ into Eq. \eqref{eq:final model},
we achieve the optimization problem w.r.t. $W$: 
\begin{equation}
\begin{split}
&\mathop{\arg\min}_{W}Tr(W^TXSX^TW)-2Tr(F^TW^TX)\\
&+Tr(S^{-1}F^TF)+\eta Tr(W^TW)+\rho Tr(W^TX_tL_tX_t^TW)\\
\hookrightarrow
&\mathop{\arg\min}_{W}\left\|(FS^{-1}-W^TX)S^{\frac{1}{2}}\right\|_F^2\\
&\qquad \quad \ +Tr(W^T(\eta I+\rho  XLX^T)W)\\
\end{split}
\label{eq:W-problem}
\end{equation}
where $L=diag(\bm{0}_{n_s\times n_s},L_t)$.
By setting the derivative of Eq. \eqref{eq:W-problem} w.r.t. $W$ to zero,
we finally obtain
\begin{equation}
W^*=(X(S+\rho L)X^T+\eta I)^{-1}XF^T
\label{eq:W-solve}
\end{equation}

\subsection{Discussion}

\subsubsection{Kernelization}

Considering the nonlinear case,
we further extend our learning model $f$ to a kernel-based nonlinear model. 
Let $f=W\phi(x)$ denotes the prediction function, where $W$ is the coefficients vector (i.e., classifier parameters), $\phi:x\to\mathcal{H}$ denotes the feature mapping function that projects the original feature vector to a Hilbert space $\mathcal{H}$ .
Then the kernel function induced by $\phi$ is defined as $K(x_i,x_j)=\langle \phi(x_i),\phi(x_j)\rangle$.
Based on the Representation theorem \cite{scholkopf2002learning},
Eq. \eqref{eq:W-problem} can be converted into the following problem:
\begin{equation}
\begin{split}
\mathop{\arg\min}_{W}&\left\|(FS^{-1}-W^TK)S^{\frac{1}{2}}\right\|_F^2\\
&+tr(W^T(\eta K+\rho  KLK)W)\end{split}
\label{eq:kernelization-problem}
\end{equation}
By setting the derivative w.r.t. $W$ to zero,
the closed-form solution of problem \eqref{eq:kernelization-problem} is given by:
\begin{equation}
W^*=((S+\rho L)K+\eta I)^{-1}F^T
\label{eq:kernelization-solve}
\end{equation}

\subsubsection{Complexity Analysis}
We analyze the computational complexity of the proposed SP-TCL in Algorithm \ref{proposed_alg}.
The construction of $M$ costs $O(n_t^2)$, which is conducted only once.
The complexity of updating $P$ is $O(|\mathcal{Y}_s|n)$,
and solving $W$ via matrix inversion occupies $O(m^3)$. 
In summary, 
the overall complexity of Algorithm \ref{proposed_alg} is $O(n_t^2+T_oT_i(|\mathcal{Y}_s|n+m^3))$,
where $T_o$ is the length
 of self-pace learning and $T_i$ is the number of the inner iterations $T_i$ for EM-like optimization scheme.
It is clearly that our method can scale to large source datasets, on which the computational complexity of SP-TCL is $O(n)$.

\section{Experiments} \label{sec:experiments}
\captionsetup[table]{font=sc,justification=centering,labelsep=space,labelfont=footnotesize}

\begin{figure}[t]
	\centering
	\includegraphics[width=3.3in]{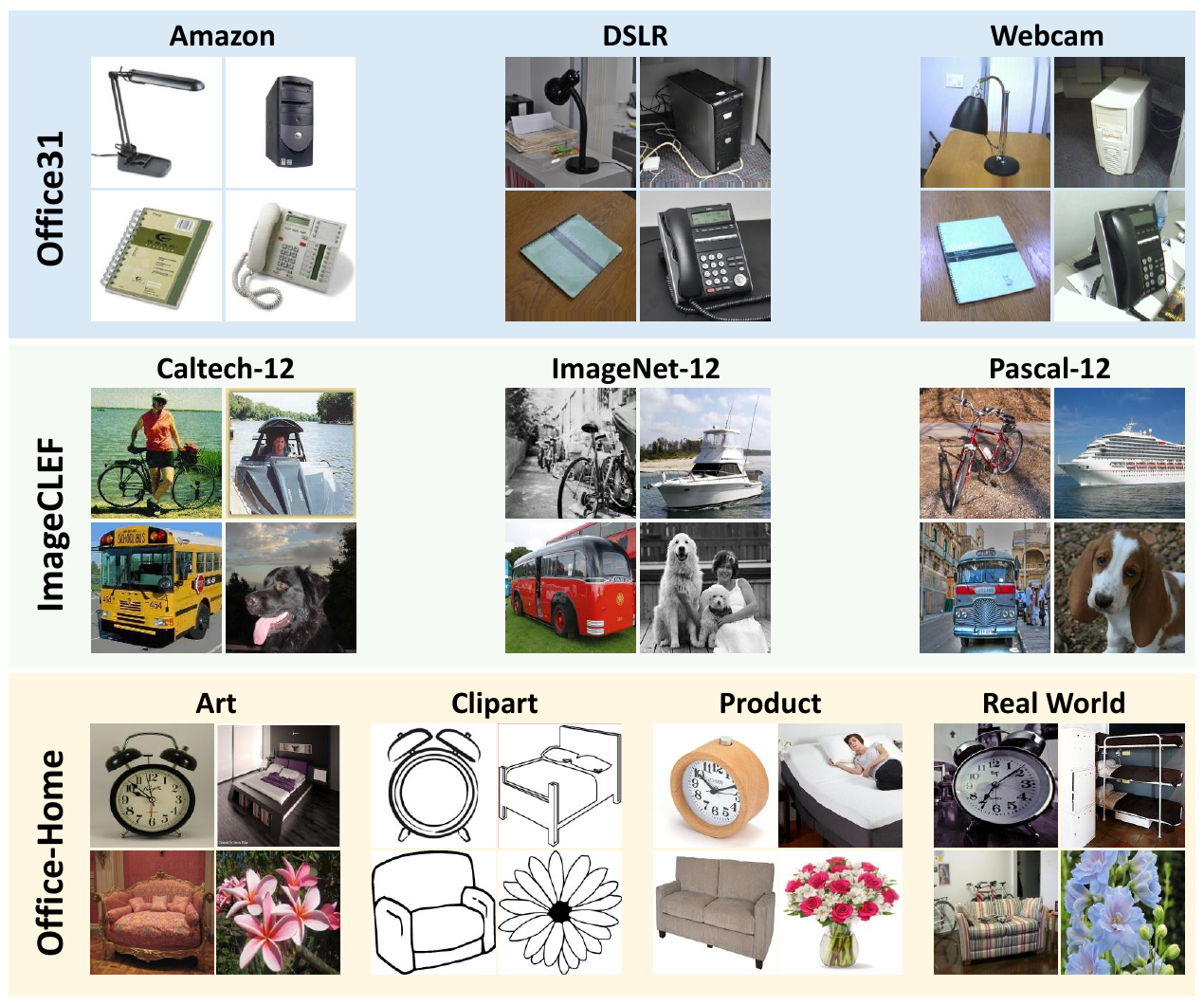}
	\captionsetup{font={footnotesize}}
	\caption{Image examples of Office31, ImageCLEF and Office-Home datasets.}
	\label{fig: dataset show}
\end{figure}

In this section, 
we conduct extensive experiments to evaluate the superiority and effectiveness of the proposed approach on three widely used benchmark datasets: 
Office31 \cite{cao2018partial}, ImageCLEF-DA and Office-Home \cite{venkateswara2017deep}. 

\begin{table*}[t]
	\centering
	\caption{\footnotesize\\Classification results (\%) for \textbf{weakly-supervised partial} domain adaptation on Office-Home with $p_{noise}=40\%$. }
	\label{tab:WSPDA on Office-Home}
	\tabcolsep4pt
	\begin{tabular}{ c c c c c c c c c c c c c c}
		\hline
		Methods & Ar$\to$Cl & Ar$\to$Pr & Ar$\to$Rw & Cl$\to$Ar & Cl$\to$Pr & Cl$\to$Rw & Pr$\to$Ar & Pr$\to$Cl & Pr$\to$Rw & Rw$\to$Ar & Rw$\to$Cl & Rw$\to$Pr & Avg.\\
		\hline
		1NN & 17.1&29.3&37.5&24.9&27.1&33.5&26.1&21.2&37.9&31.7&21.4&37.3&28.8 \\
		LapRLS & 43.2&59.6&71.5&54.2&56.8&65.5&54.4&40.5&71.6&64.8&48.3&73.7&58.7 \\
		JDA & 44.7&57.8&66.1&52.9&55.4&62.7&50.6&39.3&65.0&61.4&44.4&66.9&55.6 \\
		JGSA & 40.3 & 49.1& 64.2 & 46.9 & 48.2 & 56.8 & 42.7 & 38.3 & 60.1 & 59.7 & 43.5 & 62.7 & 51.1\\
		ResNet & 29.9&53.4&47.7&33.8&46.4&35.0&41.2&27.4&45.5&47.8&30.2&64.7&41.9 \\
		PADA&35.0&41.1&52.1&40.9&27.0&59.8&45.1&29.9&60.8&64.5&48.9&70.9&48.0 \\
		SAN&44.2&53.0&69.8&49.1&46.7&61.6&53.5&39.1&70.9&66.1&29.9&72.2&54.7\\
		TCL&39.1&54.4&65.1&50.7&44.0&58.0&60.9&38.0&72.2&62.6&40.4&68.7&54.5\\
		DRCN & 36.5 & 60.3 & 70.8 & 43.3 & 45.5 & 54.2 & 48.8 & 33.1 & 68.5 & 61.1 & 39.1 & 65.2 & 52.2\\
		\hline
		SP-TCL&\textbf{46.3}&\textbf{74.7}&79.1&\textbf{64.1}&\textbf{68.5}&\textbf{77.3}&\textbf{68.3}&43.8&\textbf{80.9}&70.6&46.7&\textbf{82.8}&\textbf{66.9}\\
		SP-KTCL & 45.5 & 71.3 & \textbf{82.8} & 62.3 & 65.2 & 76.4 & 63.9 & \textbf{46.4} & \textbf{80.9} & \textbf{72.0} & \textbf{53.3} & 81.8 & 66.8\\
		\hline
	\end{tabular}
\end{table*}

\begin{table*}[t]
	\centering
	\caption{\footnotesize\\Classification results (\%) for \textbf{weakly-supervised partial} domain adaptation on Office31 and imageCLEF with $p_{noise}=40\%$. }
	\label{tab:WSPDA on Office31}
	\tabcolsep6pt
	\begin{tabular}{ c c c c c c c  c| c c c c c c c }
		\hline
		\multirow{2}{*}{Methods}&\multicolumn{7}{c|}{Office31}&\multicolumn{7}{c}{ImageCLEF}\\
		\cline{2-15}
		& A$\to$D & A$\to$W & D$\to$A & D$\to$W & W$\to$A & W$\to$D& Avg.& C$\to$I & C$\to$P & I$\to$C & I$\to$P & P$\to$C & P$\to$I & Avg.\\
		\hline
		1NN & 51.6 & 47.5 & 52.3 & 63.6 & 50.7 & 63.9 & 54.9 & 52.9 & 48.5 & 60.9 & 45.3 & 40.3 & 50.4 & 49.7\\
		LapRLS & 84.3 & 80.5 & 78.8 & 92.4 & 86.3 & 99.2 & 86.9 & 83.1 & 70.3 & 92.9 & 77.2 & 85.3 & 81.6 & 81.7\\
		JDA & 86.8 & 81.9 & 67.9 & 87.0 & 85.8 & 97.7 & 84.5 & 70.8& 63.7 & 89.3 & 65.1 & 86.0 & 70.5 & 74.2\\
		JGSA& 69.2 & 70.6 & 56.3 & 77.0 & 77.9 & 84.3 & 72.5 & 66.7 & 60.7 & 71.3 & 63.1 & 54.0 & 73.5 & 64.9\\
		ResNet & 52.7 & 64.1 & 69.8 & 79.2 & 70.5 & 74.4 & 68.5 & 64.3 & 52.1 & 88.9 & 65.5 & 75.7 & 66.1 & 68.8\\
		PADA & 75.8 & 75.4 & 77.3 & 85.9 & 81.1 & 92.4 & 81.3 & 80.5 & 65.5 & 93.6 & 74.0 & 85.7 & 80.7 & 80.0\\
		SAN & 85.6 & 83.6 & 83.4 & 87.7 & 87.2 & 93.4 & 86.8 & 84.3 & 63.2 & 93.7 & 73.5 & 85.3 & 80.3 & 80.0\\
		TCL & 85.1 & 88.3 & 83.7 & 82.7 & 88.8 & 94.9 & 87.2 & 80.1 & 64.0 & 94.8 & 73.2 & 86.8 & 83.2 & 80.4\\
		DRCN & 87.0 & 81.5 & 77.9 & 91.5 & 79.3 & 97.9 & 85.9 & 81.9 & 66.8 & 93.3 & 74.7 & 86.5 & 82.8 & 81.0\\
		\hline
		SP-TCL& \textbf{94.3} & 86.8 & \textbf{88.8} & \textbf{96.2} & \textbf{95.6} & 99.4 & 93.5& 90.3 & \textbf{86.1} & \textbf{99.6} & \textbf{86.0} & \textbf{98.5} & \textbf{93.1} & \textbf{92.3}\\
		SP-KTCL & \textbf{94.3}&\textbf{95.1}&86.0&95.0&92.8&\textbf{100}&\textbf{93.9}& \textbf{91.2}&82.5&99.6&85.2&98.4&91.9&91.5\\
		\hline
	\end{tabular}
\end{table*}

\subsection{Dataset Description}
We visualize some sample images per dataset used in our experiments in Fig. \ref{fig: dataset show}.
The detailed information about these datasets is described as follows.

\textbf{Office31} \cite{Gong2015Geodesic,cao2018partial} is the most widely used benchmark dataset for visual domain adaptation,
which consists of 4110 images with 31 categories in 3 distinct domains:
1) Amazon (\textbf{A}, images downloaded from online merchants),
2) DSLR (\textbf{D}, high-resolution images taken digital SLR cameras),
3) Webcam (\textbf{W}, low-resolution images taken by web cameras).
For partial domain adaptation scenario,
following similar settings of \cite{cao2018partial} and \cite{cao2019learning},
we select images belonging to the 10 categories shared by Office-31 and Caltech-256 \cite{griffin2007caltech} in each domain of Office-31 as new target domains.
We evaluate all comparison methods on six cross-domain classification tasks:
\textbf{A$\to$D}, \textbf{A$\to$W}, ... , \textbf{W$\to$D}.

\textbf{ImageCLEF} dataset \cite{long2018conditional} consists of three distinct object domains:
1) Caltech-256 (\textbf{C}) \cite{griffin2007caltech}, 2) ImageNet ILSVRC 2012 (\textbf{I}) and 3) Pascal VOC 2012 (\textbf{P}). 
In our experiments,
images belonging to the 12 common categories that are shared by all these domains are selected for evaluation. 
Specifically,
we use the first 5 categories (in alphabetical order) as target categories for partial domain adaptation tasks.
By considering all combinations of different domains,
we finally built six cross-domain classification tasks:
\textbf{C$\to$I}, \textbf{C$\to$P}, ... , \textbf{P$\to$I}.

\textbf{Office-Home} \cite{venkateswara2017deep} is a more large and challenging dataset for domain adaptation,
which contains around 15500 images from 65 everyday object categories in 4 distinct domains:
1) Art (\textbf{Ar}, paintings, sketches and/or artistic depictions),
2) Clipart (\textbf{Cl}, clipart images),
3) Product (\textbf{Pr}, images without background),
4) Real-World (\textbf{Rw}, product images captured with a camera).
For partial domain adaptation scenarios,
to be consistent with protocols in \cite{cao2019learning},
we select images belonging to the first 25 categories in alphabetical order from Office-Home to form the new target domain.
Hence, we finally build $12$ cross-domain classification tasks:
\textbf{Ar$\to$Cl}, \textbf{Ar$\to$Pr}, ... , \textbf{Rw$\to$Pr}.

\begin{table*}[t]
	\centering
	\caption{\footnotesize\\Classification results (\%) for \textbf{partial} domain adaptation on Office-Home. }
	\label{tab:PDA on Office-Home}
	\tabcolsep4pt
	\begin{tabular}{ c c c c c c c c c c c c c c }
		\hline
		Methods & Ar$\to$Cl & Ar$\to$Pr & Ar$\to$Rw & Cl$\to$Ar & Cl$\to$Pr & Cl$\to$Rw & Pr$\to$Ar & Pr$\to$Cl & Pr$\to$Rw & Rw$\to$Ar & Rw$\to$Cl & Rw$\to$Pr & Avg.\\
		\hline
		1NN & 43.8&62.7&73.5&52.6&55.6&62.7&56.8&42.7&73.8&64.7&47.5&73.9&59.2\\
		LapRLS & 54.0&70.5&80.1&61.3&62.2&71.2&62.9&49.5&79.0&72.4&53.7&79.2&66.3\\
		JDA & 50.9&69.1&76.9&59.3&60.2&67.5&59.0&47.0&76.9&68.3&51.5&76.0&63.6\\
		JGSA& 48.5 & 58.2 & 67.3 & 56.3 & 55.8 & 67.9 & 58.3& 49.1 & 75.4 & 65.4 & 49.5 & 72.0 & 60.3\\
		ResNet & 38.6 & 60.8 & 75.2 & 39.9 & 48.1 & 52.9 & 49.7 & 30.9 & 70.8 & 65.4 & 41.8 & 70.4 & 53.7\\
		PADA & 52.0&67.0&78.7&52.2&53.8&59.0&52.6&43.2&78.8&73.7&56.6&77.1&62.1\\
		SAN & 44.4&68.7&74.6&67.5&65.0&77.8&59.8&44.7&80.1&72.2&50.2&78.7&65.3\\
		ETN & 59.2&77.0&79.5&62.9&65.7&75.0&68.3&55.4&84.4&75.7&57.7&84.5&70.5\\
		DPADA & 56.5 & 77.6 & 80.3 & 65.7 & 71.5 & 77.3 & 66.5 & 56.0 & 85.7 & 77.0 & 60.8 & 84.8 & 71.6\\
		DRCN & 54.0 & 76.4 & 83.0 & 62.1 & 64.5 & 71.0 & 70.8 & 49.8 & 80.5 & 77.5 & 59.1 & 79.9 & 69.0\\
		RTNet & \textbf{62.7} & 79.3 & 81.2 & 65.1 & 68.4 & 76.5 & 70.8 & 55.3 & 85.2 & 76.9 & 59.1 & 83.4 & 72.0\\
		AGCN & 56.4 & 77.3 & 85.1 & \textbf{74.2} & 73.8 & 81.1 & 70.8 & 51.5 & 84.5 & \textbf{79.0} & 56.8 & 83.4 & 72.8\\
		\hline
		SP-TCL & 60.0 & \textbf{86.6} & \textbf{90.9} & 69.4 & 72.0 & 82.1 &69.4 & \textbf{59.8} &85.9 & 78.4 & 58.8 & \textbf{86.1} & 74.9\\
		SP-KTCL & 61.7 & 85.0 & 90.2 & 67.7 & \textbf{75.9} & \textbf{82.8} & \textbf{71.0} & 57.7 & \textbf{86.9} & 76.8 & \textbf{62.2} & 84.7 & \textbf{75.2}\\
		\hline
	\end{tabular}
\end{table*}

\begin{table}[t]
	\centering
	\caption{\footnotesize\\Classification results (\%) for \textbf{partial} domain adaptation on Office31. }
	\label{tab:PDA on Office31}
	\tabcolsep3.5pt
	\begin{tabular}{ c c c c c c c c }
		\hline
		Methods & A$\to$D & A$\to$W & D$\to$A & D$\to$W & W$\to$A & W$\to$D& Avg.\\
		\hline
		1NN & 84.1 & 72.5 & 87.1 & 97.3 & 87.9 & 99.4 & 88.0\\
		LapRLS & 85.4 & 77.6 & 89.8 & 99.7 & 91.9 & \textbf{100.0} & 90.7\\
		JDA & 81.5 & 78.6 & 78.0 & 84.1 & 91.2 & 95.5 & 84.8\\
		JGSA & 80.3 & 64.4 & 74.4 & 86.1 & 86.2 & 94.9 & 81.1\\
		ResNet &83.4 & 75.6 & 83.9 & 96.3 & 85.0 & 98.1 & 87.1\\
		PADA & 82.2 & 86.5 & 92.7 & 99.3 & 95.4 & \textbf{100.0} & 92.7\\
		SAN & 94.3 & 93.9 & 94.2 & 99.3 & 88.7 & 99.4 & 95.0\\
		ETN & 95.0 & 94.5 & \textbf{96.2} & \textbf{100.0} & 94.6 & \textbf{100.0} & 96.7\\
		DAPDA & 92.2 & 95.1 & 95.1 & \textbf{100.0} & \textbf{97.4} & \textbf{100.0} & 96.6\\
		DRCN & 86.0 & 88.5 & 95.6 & \textbf{100.0} & 95.8 & \textbf{100.0} & 94.3\\
		RTNet & 97.8 & 95.1 & 93.9 & \textbf{100.0} & 94.1 & \textbf{100.0} & 96.8\\
		AGCN & 94.3 & \textbf{97.3} & 95.7 & \textbf{100.0} & 95.7 & \textbf{100.0} & \textbf{97.2}\\
		\hline
		SP-TCL&\textbf{98.7}&94.6&95.5&99.7&95.5&99.4&\textbf{97.2}\\
		SP-KTCL& 96.8 & 84.1 & 95.5 & \textbf{100.0} & 95.4 & \textbf{100.0} & 95.3\\
		\hline
	\end{tabular}
\end{table}

\subsection{Comparison Methods and Experiment Setting}
To extensively verify the effectiveness of the proposed method,
we compare SP-TCL with 3 baselines and several state-of-the-art domain adaptation approaches. 
We adopt 1NN, LapRLS \cite{belkin2006manifold} and ResNet-50 \cite{he2016deep} as the baselines.
20 traditional or generalized domain adaptation approaches include
Joint Domain Adaptation (JDA) \cite{Long2014Transfer},
Joint Geometric and Statistical Alignment (JGSA) \cite{zhang2017joint},
Minimum Centroid Shift (MCS) \cite{liang2019distant},
Locality Preserving Joint Transfer (LPJT) \cite{li2019locality},
Graph Embedding Framework based on LDA (GEF-LDA) \cite{8767033},
Joint Adaptation Networks (JAN) \cite{long2017deep},
Conditional Domain Adversarial Networks (CDAN) \cite{long2018conditional},
Domain-specific Whitening Transform (DWT) \cite{roy2019unsupervised},
Partial Adversarial Domain Adaptation (PADA) \cite{cao2018partial_PADA},
Selective Adversarial Networks (SAN) \cite{cao2018partial},
Example Transfer Network (ETN) \cite{cao2019learning},
Dual Alignment for Partial Domain Adaptation (DAPDA) \cite{li2020dual},
Transferable Curriculum Learning (TCL) \cite{shu2019transferable},
Deep Residual Correction Network (DRCN) \cite{li2020deep},
Reinforced Transfer Network (RTNet) \cite{chen2020selective},
Dual-Representation AutoEncoder (DRAE) \cite{9314101},
Adaptive Graph Adversarial Networks (AGAN) \cite{kim2021adaptive},
Self-Paced Collaborative and Adversarial Network (SPCAN) \cite{8943120},
Spectral UDA (SUDA) \cite{zhang2022spectral},
Category Contrast technique (CaCo) \cite{huang2022category}.

Following standard protocols for unsupervised domain adaptation, 
we take all labeled source data and unlabeled target data for training. 
For a fair comparison, 
we directly extract deep features using ResNet-50 pre-trained on ImageNet dataset for shallow methods if the experiment is performed under the weakly-supervised scenarios,
otherwise deep features of ResNet-50 fine-tuned on labeled source domain are used for shallow methods.
For label corruption,
following the protocol in \cite{shu2019transferable},
we modify the label of each image evenly to a random class with probability $p_{noise}$.
For the proposed SP-TCL,
we empirically fix the regularization parameter $\rho=1$ in all experiments,
while determining the optimal value of free parameters $\eta$ and $r$ through grid-search.
We will give the parameter sensitivity test for SP-TCL in Section \ref{exp:para_analysis},
which indicates that SP-TCL can always achieve promising performance under reasonable parameter settings.
The RBF kernel is adopted to evaluate the kernel vision of our method (denoted as SP-KTCL).
For all comparison methods,
we use the experimental results reported in the published papers if the experiment settings are identical to ours, 
otherwise, we conduct experiments using the source implementations provided by the authors.
Particularly,
we replace the 1NN classifier with SVM classifier for JDA and JGSA.
We report the average performance over three independent trials per experiment under the weakly-supervised scenarios. 

\subsection{Performance Evaluation}
We evaluate the performance of SP-TCL by comparing with several competitive state-of-the-art domain adaptation methods on the following generalized or traditional domain adaptation tasks.
\subsubsection{Experimental results on WS-PDA tasks}

The classification results of weakly-supervised partial domain adaptation tasks on all datasets are shown in Tables \ref{tab:WSPDA on Office-Home}-\ref{tab:WSPDA on Office31},
from which we can clearly observe that the proposed method significantly outperforms all comparison methods by remarkable margins in all cross-domain tasks.
This phenomenon demonstrates the advantage of SP-TCL in knowledge discovery and knowledge adaptation under the self-paced learning fashion.
From these results, we further have some interesting observations.

Firstly, 1NN classifier achieves the worst results on all datasets,
as 1NN will be seriously affected by noisy source labels.
LapRLS obtains much performance improvement over 1NN,
because the $\ell_2$ regularization in LapRLS can prevent over-fitting on the noisy source data to a certain extent,
and the Laplacian regularizer can further enhance the performance by utilizing the local structure information of data.
However, SP-TCL beats LapRLS by an average accuracy improvement of $8.2\%$, $6.6\%$ and $10.6\%$ on Office-Home, Office31 and ImageCLEF datasets, respectively.
This verifies the superiority of the proposed method for WS-PDA tasks.

Secondly, JDA and JGSA show relatively poor performance compared to SP-TCL, 
which can be attributed to that these two approaches will suffer from severe negative transfer caused by noisy/outlier examples. 
This phenomenon indicates that performing distribution adaptation is not suitable for knowledge transfer when the source domain and target domain hold different label spaces.
\begin{table}[t]
	\centering
	\caption{\footnotesize\\Classification results (\%) for \textbf{weakly-supervised} domain adaptation on Office31 with $p_{noise}=40\%$. }
	\label{tab:WS-DA on Office31}
	\tabcolsep3.5pt
	\begin{tabular}{ c c c c c c c c }
		\hline
		Methods & A$\to$D & A$\to$W & D$\to$A & D$\to$W & W$\to$A & W$\to$D& Avg.\\
		\hline
		1NN & 44.5 & 44.4 & 38.2 & 60.3 & 37.0 & 57.4 & 47.0\\
		LapRLS & 75.0 & 71.0 & 62.3 & 90.0 & 59.7 &92.8 &75.1\\
		JDA & 64.2 & 65.3 & 57.9 & 85.5 & 55.5 & 86.2 & 69.1\\
		JGSA & 38.9 & 40.9 & 50.9 & 73.1 & 45.5 & 74.5 & 54.0\\
		ResNet & 47.1 & 47.2& 31.0&58.8&33.0&68.0&47.5\\
		TCL & 83.3 & 82.0 & 60.5 & 77.2 & 65.7 & 90.8 & 76.6\\
		\hline
		SP-TCL&\textbf{86.6}&85.8&69.8&93.8&\textbf{67.6}&97.7&\textbf{83.6}\\
		SP-KTCL & 86.0 & \textbf{86.3} & \textbf{70.2} & \textbf{95.8} & 62.8 & \textbf{97.8} & 83.1\\
		\hline
	\end{tabular}
\end{table}

\begin{table*}[t]
	\centering
	\caption{\footnotesize\\Classification results (\%) for \textbf{traditional} domain adaptation on Office-Home. }
	\label{tab:DA on Office-Home}
	\tabcolsep4pt
	\begin{tabular}{c c c c c c c c c c c c c c }
		\hline
		Methods&Ar$\to$Cl&Ar$\to$Pr&Ar$\to$Rw&Cl$\to$Ar&Cl$\to$Pr&Cl$\to$Rw&Pr$\to$Ar&Pr$\to$Cl&Pr$\to$Rw&Rw$\to$Ar&Rw$\to$Cl&Rw$\to$Pr&Avg.\\
		\hline
		1NN&43.2&61.2&67.8&47.1&58.9&60.9&50.1&42.3&69.9&61.6&47.7&75.2&57.2\\
		JDA&51.1&67.4&70.9&51.3&64.1&64.3&54.4&47.7&73.4&64.6&53.7&78.7&61.8\\
		JGSA&51.1&71.0&74.3&53.1&68.8&68.8&56.4&48.0&76.3&65.1&52.9&78.9&63.7\\
		MCS&55.9&73.8&79.0&57.5&69.9&71.3&58.4&50.3&\textbf{78.2}&65.9&53.2&\textbf{82.2}&66.3\\
		LPJT&52.9&73.3&75.6&54.8&66.7&69.4&54.8&49.9&75.8&65.8&55.2&80.5&64.6\\
		GEF-LDA&52.1&69.4&73.6&53.9&66.5&68.2&56.8&49.2&74.5&67.5&55.0&80.8&64.0\\
		ResNet&34.9&50.0&58.0&37.4&41.9&46.2&38.5&31.2&60.4&53.9&41.2&59.9&46.1\\
		JAN&45.9&61.2&68.9&50.4&59.7&61.0&45.8&43.4&70.3&63.9&52.4&76.8&58.3\\
		CDAN&49.0&69.3&74.5&54.4&66.0&68.4&55.6&48.3&75.9&68.4&55.4&80.5&63.8\\
		DWT&50.3&72.1&77.0&59.6&69.3&70.2&58.3&48.1&77.3&69.3&53.6&82.0&65.6\\
		DRAE&53.4&73.7&76.9&55.8&69.9&69.5&55.4&48.1&77.4&65.6&53.7&80.2&65.0\\
		DRCN & 50.6 & 72.4 & 76.8 & \textbf{61.9} & 69.5 & 71.3 & 60.4 & 48.6 & 76.8 & 72.9 & 56.1 & 81.4 & 66.6\\
		SP-TCL& \textbf{56.4} & \textbf{76.9} & \textbf{78.6} & 61.0 & \textbf{73.6} & \textbf{72.3} & \textbf{61.2} & \textbf{54.5} & 77.5 & \textbf{79.5} & \textbf{57.8} & 81.5 & \textbf{68.5}\\
		\hline
	\end{tabular}
\end{table*}

\begin{table}[t]
	\centering
	\caption{\footnotesize\\Classification results (\%) for \textbf{traditional} domain adaptation on ImageCLEF. }
	\label{tab:DA on ImageCLEF}
	\tabcolsep5pt
	\begin{tabular}{c c c c c c c c}
		\hline
		Methods & C$\to$I & C$\to$P & I$\to$C & I$\to$P & P$\to$C & P$\to$I & Avg.\\
		\hline
		1NN&84.0&67.8&90.8&75.2&82.2&79.2&79.9\\
		JDA&91.3&76.5&95.0&77.5&84.3&78.3&83.8\\
		JGSA&92.2&76.5&94.5&78.0&88.2&85.0&85.7\\
		LPJT&91.3&77.0&94.7&77.8&91.0&85.3&86.2\\
		ResNet&78.0&65.5&91.5&74.8&91.2&83.9&81.7\\
		JAN&89.5&74.2&94.7&76.8&91.7&88.0&85.8\\
		CDAN&90.5&74.5&\textbf{97.0}&76.7&93.5&90.6&87.1\\
		DWT&87.5&73.4&94.3&77.7&94.5&89.7&86.2\\
		SPCAN&\textbf{92.9}&\textbf{79.4}&95.5&\textbf{79.0}&91.3&91.1&88.2\\
		SP-TCL & 92.8 & 78.3 & 96.7 & 77.8 & \textbf{95.5} & \textbf{93.2} & \textbf{89.1}\\
		\hline
	\end{tabular}
\end{table}

Thirdly, we observe that the performance of partial domain adaptation methods, i.e., PADA , SAN and DRCN, are relatively poor when compared to SP-TCL. 
The reason behind this is that their weighting mechanism can mitigate negative transfer caused by outlier classes, 
however, still suffer from over-fitting caused by noisy source examples. 
The noisy source examples will severely deteriorate the learning procedure of the domain adversarial network, 
thereby incurring a huge performance drop. 
Moreover, TCL, which is a weakly-supervised domain adaptation algorithm,
shows inferior performance to SP-TCL.
Even though TCL can handle noisy source labels in model training,
it still suffers from negative transfer caused by outlier classes.

Finally, SP-TCL achieves the best results among all comparison approaches in all cross-domain tasks.
To be specific, SP-TCL gains the performance improvement of $8.2\%$ , $6.3\%$ and $10.6\%$ over the best competitor on the three datasets, respectively.
The results prove that SP-TCL is effective for WS-PDA by knowledge discovery via the prudent loss function, 
as well as knowledge adaptation via the self-pace learning strategy and manifold regularization.
Note that the non-linear version of our method, i.e., SP-KTCL, obtains similar performance to SP-TCL.
In summary,
by adopting deep features extracted by ResNet-50 pre-trained on ImageNet without fine-tuned on labeled source domain,
SP-TCL can still achieve superior performance compared with these deep learning models, 
which verifies the effectiveness of our proposed method.

\subsubsection{Experimental results on PDA or WS-DA tasks}

We further investigate the effectiveness of the proposed method under the scenario of partial domain adaptation or weakly-supervised domain adaptation respectively.
Table \ref{tab:PDA on Office-Home} and \ref{tab:PDA on Office31} summarize the results on Office-Home and Office31 under partial domain adaptation scenario,
in which SP-TCL achieves significant performance improvement over all comparison methods on Office-Home,
and shows comparable results with PDA approaches on Office31.
Specifically,
our approach obtains the highest accuracies in 9 out of 12 cross-domain tasks on Office-Home,
with an average accuracy improvement of $2.4\%$ over AGCN. 
SP-TCL also yields the highest average accuracy on Office31.
From the results, we can observe that PDA approaches usually achieve much better results than traditional UDA approaches.
For example, RTNet gains average performance improvement of $5.7\%$ and $12\%$ over JDA on Office-Home and Office31, respectively.
It is worth to note that the results of JDA and JGSA are worse than the baseline 1NN in terms of average accuracy on Office31,
indicating that irrelevant source classes will bring serious negative transfer in the distribution adaptation process.
Different from existing methods that aim at establishing knowledge transfer from the perspective of transferable features/instances, 
SP-TCL performs knowledge transfer based on the classifier adaptation from the source domain to the target domain,
which relaxes the assumption of identical label spaces and can be applied to a more generalized setting of domain adaptation. 

Then, we list the experimental results of weakly-supervised domain adaptation tasks on the Office31 dataset in Table \ref{tab:WS-DA on Office31}.
It is obvious that the proposed method notably outperforms other comparison methods in all cross-domain tasks,
and the average accuracy is $6.0\%$ higher than TCL.
The results prove that the prudent loss function adopted in SP-TCL is effective for knowledge discovery, thereby improving cross-domain classification accuracy. 
The weakly-supervised domain adaptation method, i.e., TCL, performs better than the baselines and traditional domain adaptation approaches,
due to the fact that TCL introduces the transferable curriculum learning strategy to select noiseless and transferable source examples for model training,
which can mitigate the negative transfer of noisy source examples.
JDA and JGSA show clearly poor results on WS-DA tasks,
demonstrating that the WS-DA task is a complex and difficult scenario that can not be easily handled by traditional UDA methods.

\begin{table}[t]
	\centering
	\caption{\footnotesize\\Classification results (\%) for \textbf{traditional} domain adaptation on office31. }
	\label{tab:DA on office31}
	\tabcolsep4pt
	\begin{tabular}{c c c c c c c c}
		\hline
		Methods & A$\to$D & A$\to$W & D$\to$A & D$\to$W & W$\to$A & W$\to$D& Avg.\\
		\hline
		1NN & 79.3 & 76.6 & 63.9 & 97.2 & 63.2 & 99.6 & 80.0\\
		JDA& 89.4 & 85.3 & 68.2 & 97.4 & 70.9 & 99.6 & 85.1\\
		JGSA& 86.8 & 85.4 & 69.7 & 97.6& 71.3 & 99.6 & 85.0\\
		ResNet&  68.9 &  68.4 & 62.5 & 96.7 & 60.7 & 99.3& 76.1\\
		CDAN & 89.8 & 93.1 & 71.4 & 98.2 & 68.0 & \textbf{100.0} & 86.5\\
		DRCN & 89.4 & \textbf{93.1} & 71.4 & 98.0 & 71.0 & \textbf{100.0} & 87.2\\
		SPCAN& 91.2 & 92.4 & \textbf{77.1} & \textbf{99.2} & \textbf{74.5} & \textbf{100.0} & \textbf{89.1}\\
		SUDA & 91.2 & 90.8 & 72.2 & 98.7 & 71.4 & \textbf{100.0} &  87.4\\
		CaCo & 91.7 & 89.7 & 73.1 & 98.4 & 72.8 & \textbf{100.0} & 87.6 \\
		SP-TCL & \textbf{94.0} & 89.4 & 73.6 & 98.2 & \textbf{74.5} & 99.4 & 88.2\\
		\hline
	\end{tabular}
\end{table}

\subsubsection{Experimental results on UDA tasks}
Traditional unsupervised domain adaptation is a special case of WS-PDA where the label spaces across domains are identical and $p_{noise}=0$. 
Therefore,
it is natural to extend our method to traditional UDA scenarios.
We compare the proposed method with several state-of-the-art UDA approaches on UDA tasks,
and the results on Office-Home, ImageCLEF and Offce31 datasets are illustrated in Table \ref{tab:DA on Office-Home}, \ref{tab:DA on ImageCLEF} and \ref{tab:DA on office31}, respectively.

From Table \ref{tab:DA on Office-Home}, 
we can see that SP-TCL achieves the best performance among all comparison methods.
To be specific, SP-TCL outperforms all other methods on 9 out of 12 cross-domain tasks and gains the performance improvement of $1.9\%$ over the closest competitor DRCN.
Similar patterns can also be observed in Table \ref{tab:DA on ImageCLEF},
in which SP-TCL obtains the best average result,
with an average accuracy increase of $0.9\%$ over SPCAN.
In Table \ref{tab:DA on office31}, 
SP-TCL significantly outperforms the shallow domain adaptation methods, i.e., JDA and JGSA,
and still achieve the second-best result in terms of average accuracy compared to deep domain adaptation models.
This phenomenon validates that
with the help of self-pace learning and manifold regularization, 
SP-TCL is capable of mitigating the domain shift through the proposed generalized knowledge adaptation strategy.
Thence, 
SP-TCL has the potential to address traditional unsupervised domain adaptation tasks.

\subsection{Empirical Analysis}
In this section, we will perform several empirical experiments to investigate the effectiveness of SP-TCL.
\begin{figure}[t]
	\begin{minipage}[t]{0.32\linewidth}
		\centering
		\includegraphics[width=1in]{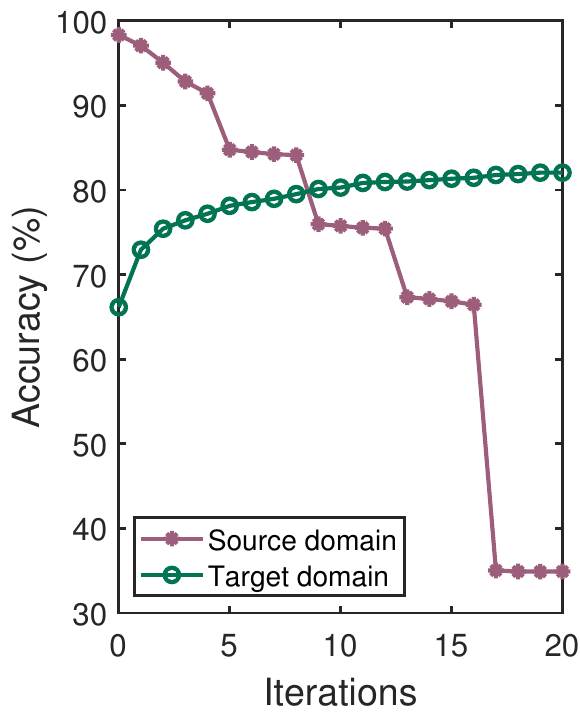}
	\end{minipage}
	\begin{minipage}[t]{0.68\linewidth}
		\centering
		\includegraphics[width=2.3in]{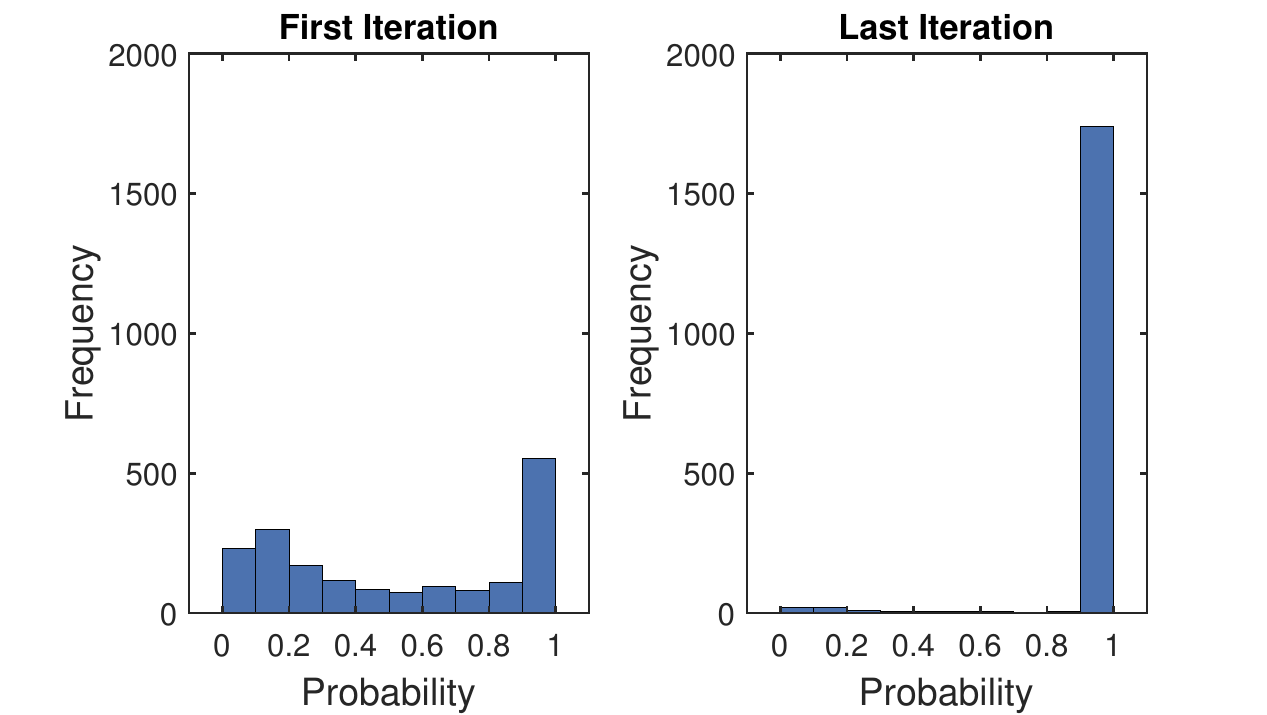}
	\end{minipage}
	
	\begin{minipage}[t]{0.32\linewidth}
		\centering
		\includegraphics[width=1in]{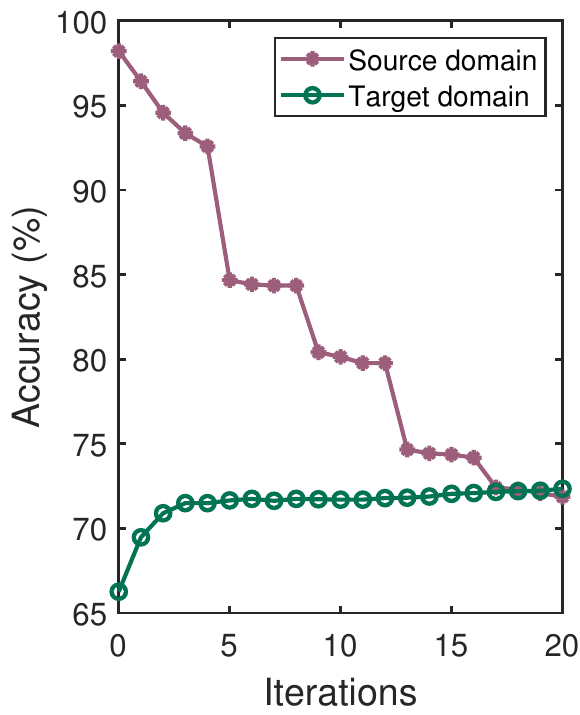}
		\centerline{(a)}
	\end{minipage}
	\begin{minipage}[t]{0.68\linewidth}
		\centering
		\includegraphics[width=2.3in]{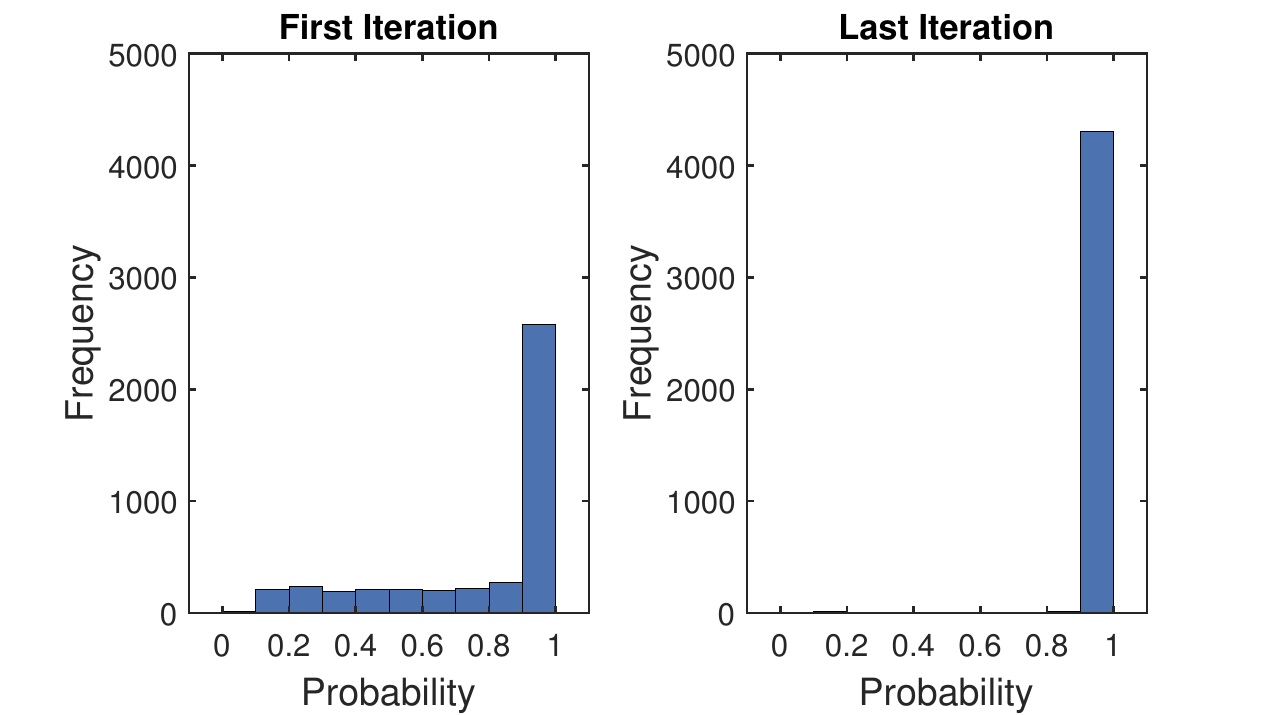}
		\centerline{(b)}
	\end{minipage}
	\captionsetup{font={footnotesize}}
	\caption{(a) Accuracies versus iterations on the task Cl$\to$Rw ; (b) Histograms of class probability distribution of Rw in the target domain on Cl$\to$Rw. 
		The top row and bottom row show the results under the scenarios of PDA and UDA, respectively.
	}
	\label{fig:Transferable study}
\end{figure}

\begin{table}[t]
	\centering
	\caption{\footnotesize\\Comparison of SP-TCL and its variant SP-TCL (w/o SPL) on ImageCLEF and Office31 datasets.}
	\label{tab:SPL study on Office31 and imageCLEF}
	\tabcolsep2.5pt
	\begin{tabular}{ c c c c c c c c }
		\hline
		\multirow{2}{*}{Methods}&\multicolumn{7}{c}{ImageCLEF}\\
		\cline{2-8}
		& C$\to$I & C$\to$P & I$\to$C & I$\to$P & P$\to$I & P$\to$C & Avg.\\
		\hline
		SP-TCL (w/o SPL) & 89.3 & 80.5 & 98.9 & 84.0 & 91.5 & 91.5 & 89.3\\
		SP-TCL & \textbf{90.3} & \textbf{86.1} & \textbf{99.6} & \textbf{86.0} & \textbf{98.5} & \textbf{93.1} & \textbf{92.3}\\
		\hline
		\hline
		\multirow{2}{*}{Methods}&\multicolumn{7}{c}{Office31}\\
		\cline{2-8}
		& A$\to$C & A$\to$D & D$\to$A & D$\to$C & W$\to$A & W$\to$D & Avg.\\
		\hline
		SP-TCL (w/o SPL) & 93.6 & 87.5 & 95.4 & \textbf{99.7} & 95.4 & \textbf{99.4} & 95.2\\
		SP-TCL & \textbf{98.7} & \textbf{94.6} & \textbf{95.5} & \textbf{99.7} & \textbf{95.5} & \textbf{99.4} & \textbf{97.2}\\
		\hline
	\end{tabular}
\end{table}
\begin{figure}[!]
	\begin{minipage}[t]{0.49\linewidth}
		\centering
		\includegraphics[width=1.5in]{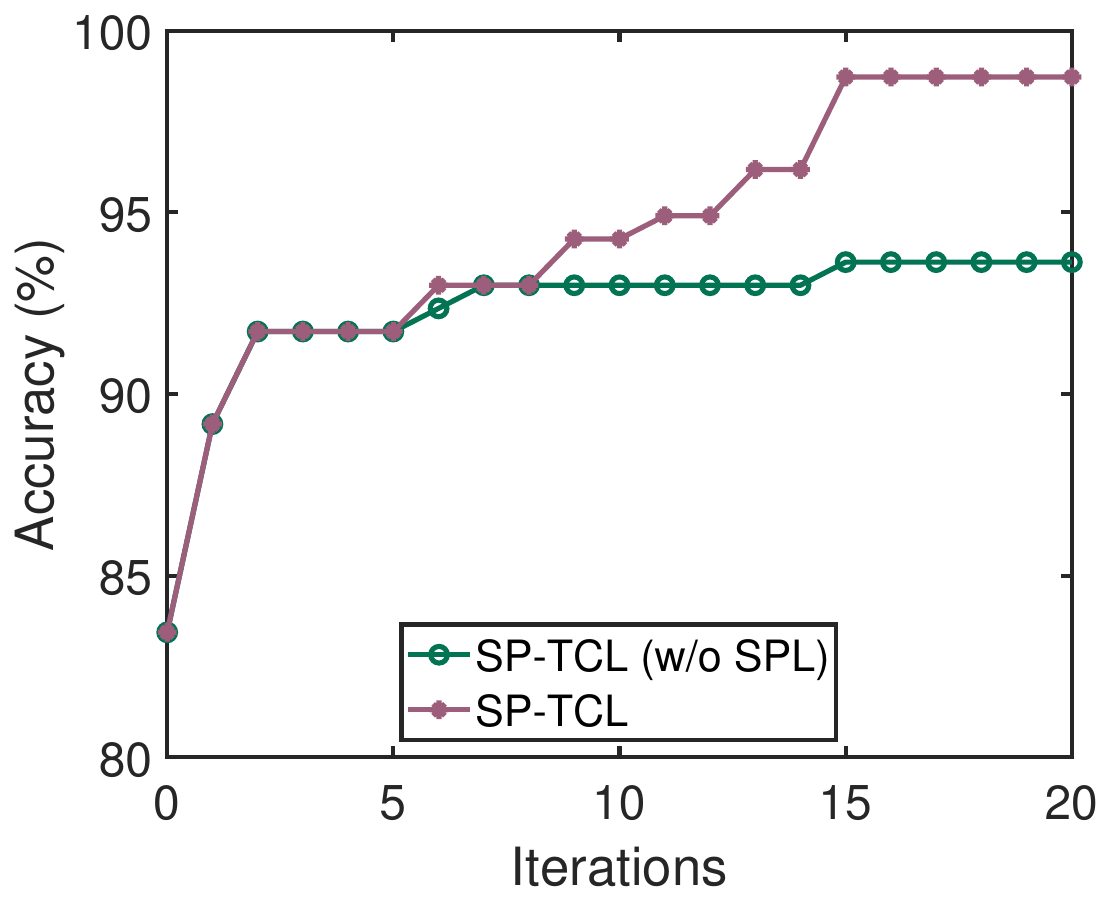}
		\centerline{(a)}
	\end{minipage}
	\begin{minipage}[t]{0.49\linewidth}
		\centering
		\includegraphics[width=1.5in]{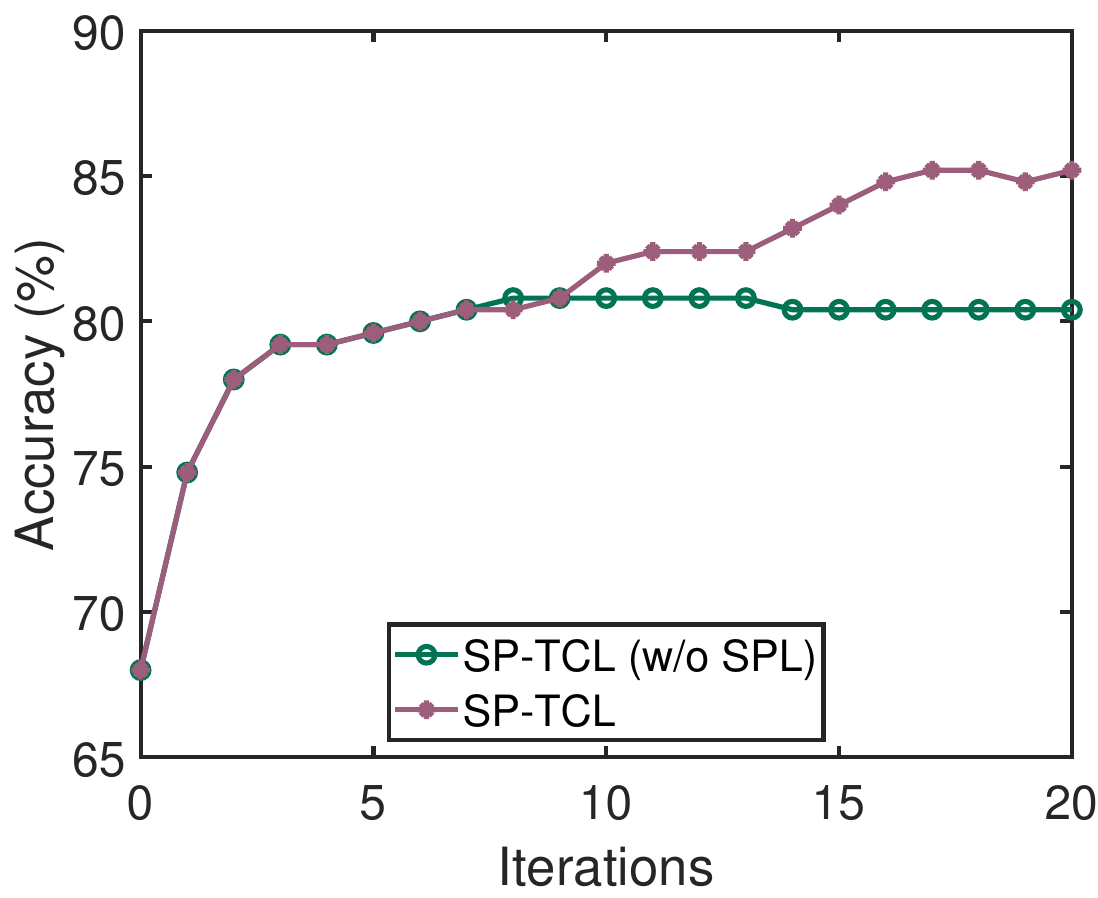}
		\centerline{(b)}
	\end{minipage}
	\captionsetup{font={footnotesize}}
	\caption{Comparison of SP-TCL and its variant SP-TCL (w/o SPL) on (a) A$\to$D and (b) C$\to$P.}
	\label{fig:SPL study}
\end{figure}

\begin{figure}[t]
	\begin{minipage}[t]{0.49\linewidth}
		\centering
		\includegraphics[width=1.5in]{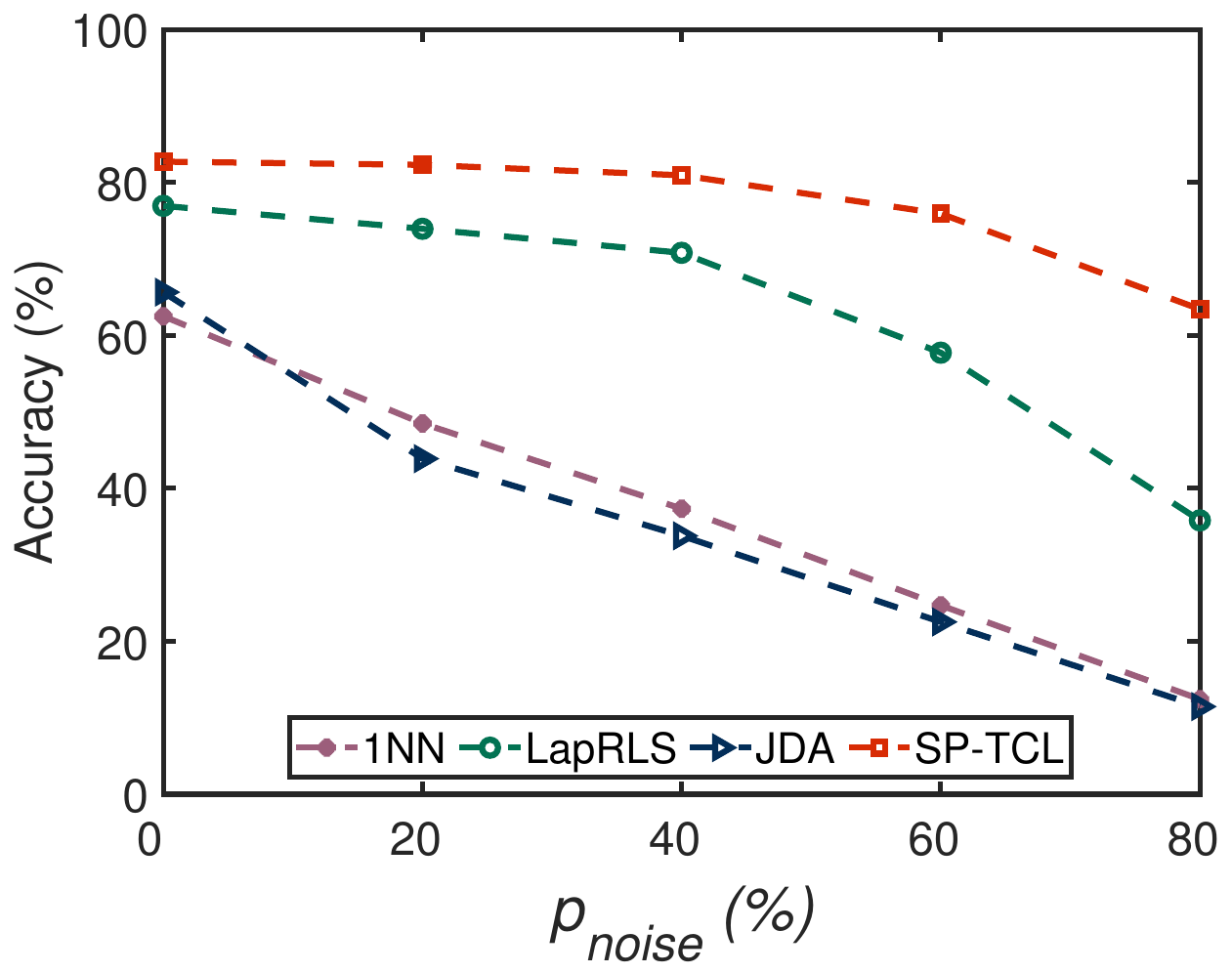}
		\centerline{(a)}
	\end{minipage}
	\begin{minipage}[t]{0.49\linewidth}
		\centering
		\includegraphics[width=1.5in]{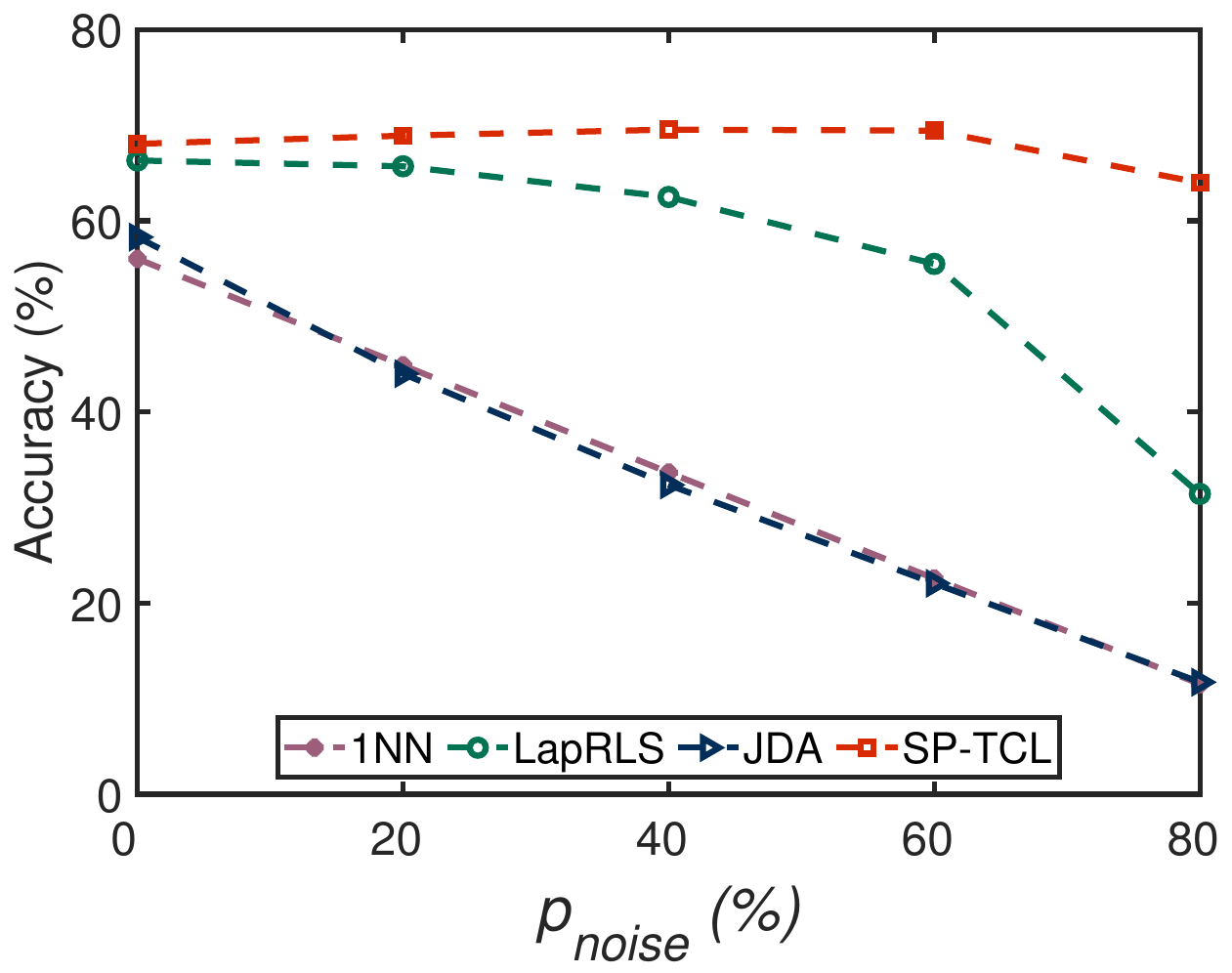}
		\centerline{(b)}
	\end{minipage}
	\captionsetup{font={footnotesize}}
	\caption{Classification accuracy w.r.t. noise levels on (a) Ar$\to$Rw and (b) Rw$\to$Ar.}
	\label{fig:noise study}
\end{figure}

\begin{figure}[!]
	\begin{minipage}[t]{0.49\linewidth}
		\centering
		\includegraphics[width=1.5in]{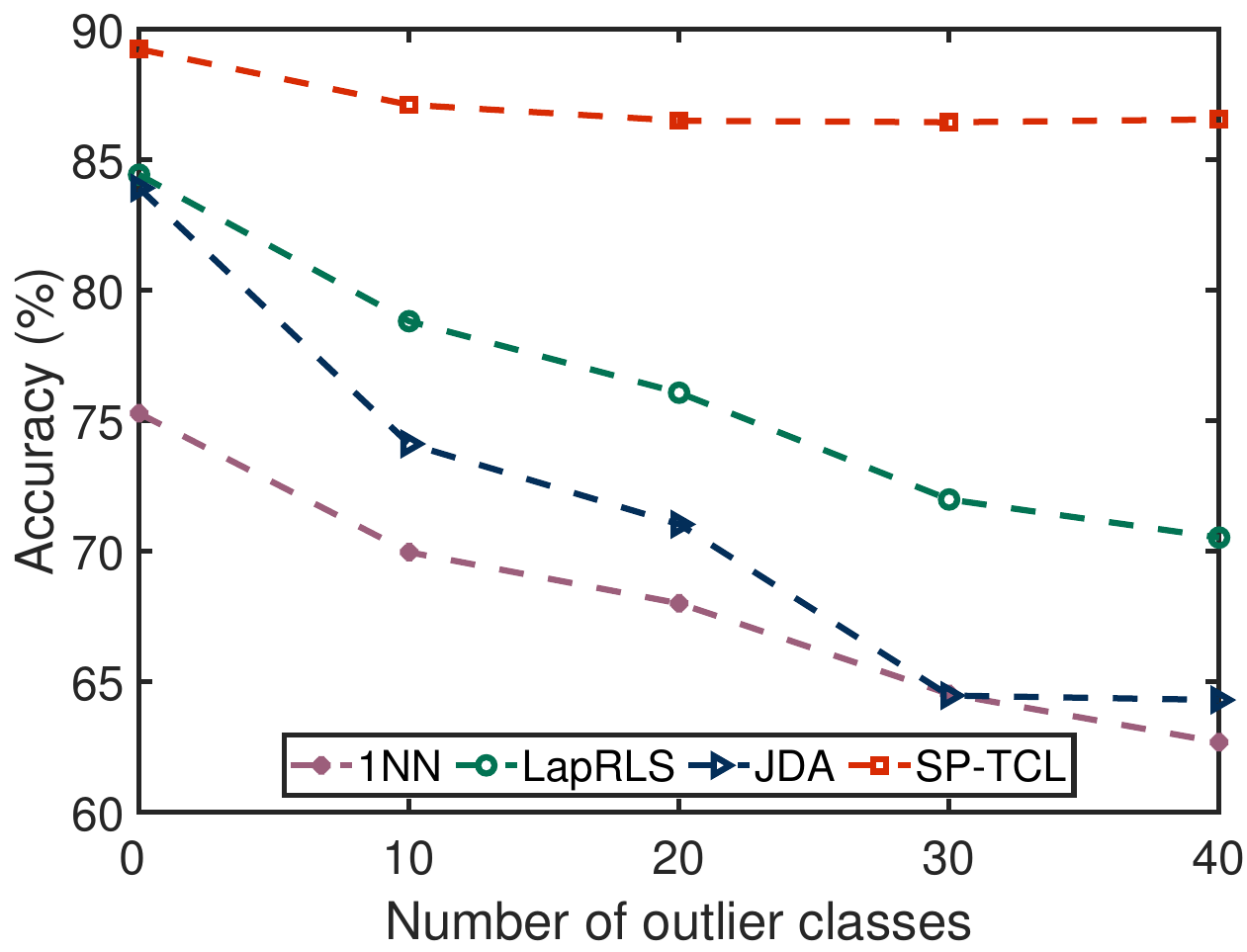}
		\centerline{(a)}
	\end{minipage}
	\begin{minipage}[t]{0.49\linewidth}
		\centering
		\includegraphics[width=1.5in]{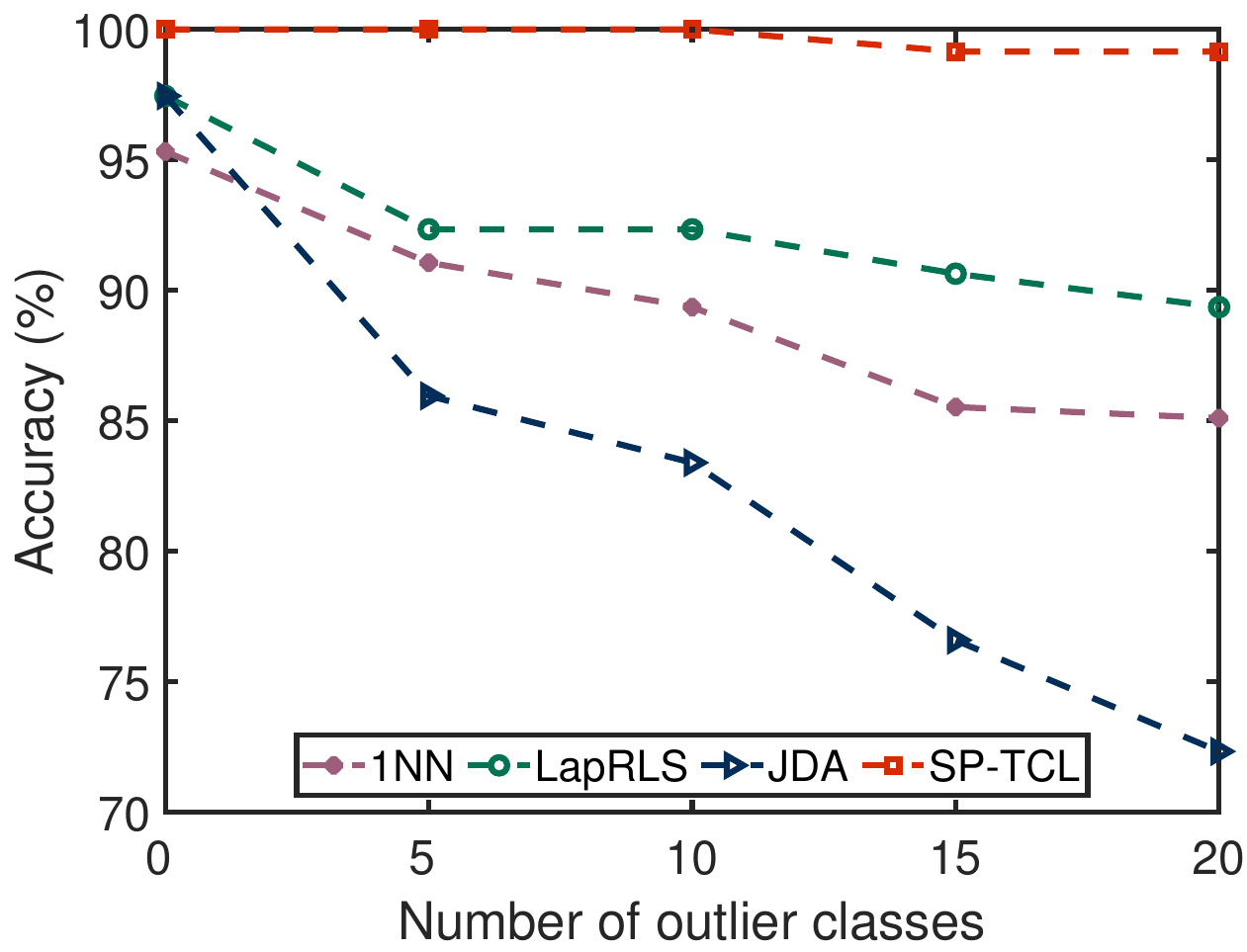}
		\centerline{(b)}
	\end{minipage}
	\captionsetup{font={footnotesize}}
	\caption{Classification accuracy w.r.t. partial levels on (a) Ar$\to$Pr and (b) A$\to$W.}
	\label{fig:partial study}
\end{figure}

\subsubsection{Transferability Study}
To verify the transferability of the classifier learned by our model, 
we perform experiments on the task Cl$\to$Rw under the scenarios of PDA and UDA,
where we can access the true source labels to calculate the accuracy of the source domain.
The classification accuracies on the source and target domains versus the iterations are shown in Fig. \ref{fig:Transferable study} (a). 
As observed, 
the best accuracy on the source domain appears in the first iteration, 
due to the fact that the classifier trained solely on the source domain is ensured to have a low source expected error. 
Such a classifier, however, will introduce a certain domain bias, resulting in a large expected error on the target domain. 
With an increasing number of iterations, 
our model iteratively excludes source examples from training 
so that the classifier can be progressively transferred to the target domain to better fit the target distribution. 
As expected, 
the accuracy gradually increases on the target domain while decreasing on the source domain. 
At the last iteration,
we completed the transition from a joint classifier to a target-preferable classifier.

We also show the distribution of the largest class probability of the target example in Fig. \ref{fig:Transferable study} (b), 
where the bar represents the number of target examples whose largest class probabilities locate in the corresponding region. 
In the beginning, 
most target examples are classified with low confidence since we are typically not certain about the labels of target examples. 
While at the end, 
almost all target examples are classified into a certain category with high confidence, 
which implies that the classifier has been successfully transferred to the target domain and fully adapted to the target distribution.

\subsubsection{SPL Regularizer Analysis}
In our model,
self-paced learning is adopted for knowledge adaptation by gradually excluding source examples from training,
through which SP-TCL can achieve the transition from a joint classifier to a target-preferable classifier.
To investigate the effectiveness of our SPL strategy,
we compare SP-TCL and its variant SP-TCL (w/o SPL), which denotes SP-TCL without SPL by selecting all source data for model training throughout the training process. 
The experimental results on ImageCLEF and Office31 datasets are illustrated in Table \ref{tab:SPL study on Office31 and imageCLEF}.
It can be observed that SP-TCL achieves much better results than SP-TCL (w/o SPL) in almost all cases,
with an increase of $3.0\%$ and $2.0\%$ in terms of average accuracy on ImageCLEF and Office31 respectively.
Note that, on the Office31 dataset, 
the performance improvement of SP-TCL is particularly significant on the tasks of A$\to$D and A$\to$W.
Perhaps the reason is that Amazon is much larger than DSLR and Webcam and the images of Amazon are without background, 
while the images of DSLR and Webcam were taken in the office environment. 
Accordingly, the domain gaps of A$\to$D and A$\to$W are larger than those of other cross-domain tasks.
In such cases,
self-paced learning would show superior performance in knowledge adaptation.

Furthermore,
to better understand the behavior of self-paced learning in the domain adaptation process,
we plot the curves of accuracy versus the iterations on the tasks of A$\to$D and C$\to$P using different variants of our model, 
and the results are shown in Fig. \ref{fig:SPL study}.
In the early phase,
they show similar performance as all source data are selected for training.
With an increasing number of iterations, 
SP-TCL iteratively excludes complex source examples from training, 
which leaves the classifier with a certain freedom to fit the target distribution, thereby continuously improving the accuracy. 
In contrast,
after a little increase, 
the curve of SP-TCL (without SPL) remains in a lower and stable state, 
which can be attributed to the fact that the optimal joint classifier is not guaranteed in the present of domain shift. 
The results in Table \ref{tab:SPL study on Office31 and imageCLEF} and Fig. \ref{fig:SPL study} indicate that 
the self-paced learning strategy can substantially enhance the knowledge adaptation process so as to learn a target-preferable classifier. 

\subsubsection{Noise Levels \& Partial Levels}
We first investigate the impact of the level of noises on two weakly-supervised partial domain adaptation tasks, i.e., Ar$\to$Rw and Rw$\to$Ar.
Fig. \ref{fig:noise study} shows that SP-TCL achieves much better performance than all the comparison methods at each noise level,
which proves the advantage of our generalized knowledge discovery strategy in dealing with noisy samples.
1NN and JDA drop rapidly with increasing levels of noise because noisy source data may severely deteriorate the source classifier and domain adaptation module.
LapRLS gains better performance improvement than 1NN and JDA because of the $\ell_2$ regularization and the Laplacian regularizer.

We further investigate the impact of outlier classes in the source domain on the proposed SP-TCL under the partial domain adaptation scenario. 
In particular, on the Office-Home dataset, 
the target domain always contains the top 25 categories in alphabetical order, while the number of categories in the source domain varies from 25 to 65 with an interval of 10.
That is, the number of outlier classes varies from 0 to 40.
On the Office31 dataset, 
the target domain consists of the top 10 categories in alphabetical order, and the number of outlier classes in the source domain varies from 0 to 20 with an interval of 5.
Fig. \ref{fig:partial study} shows the results on two randomly chosen partial domain adaptation tasks, i.e., Ar$\to$Pr and A$\to$W. 
As observed, the curves of 1NN, LapRLS and JDA drop rapidly as the number of outlier classes increases.
In contrast, based on classifier adaptation, 
SP-TCL almost stands at the same performance level, 
which strongly demonstrates its effectiveness in eliminating the negative effect of outlier classes on partial domain adaptation tasks.

\begin{figure}[t]
	\centering
	\includegraphics[width=3.5in]{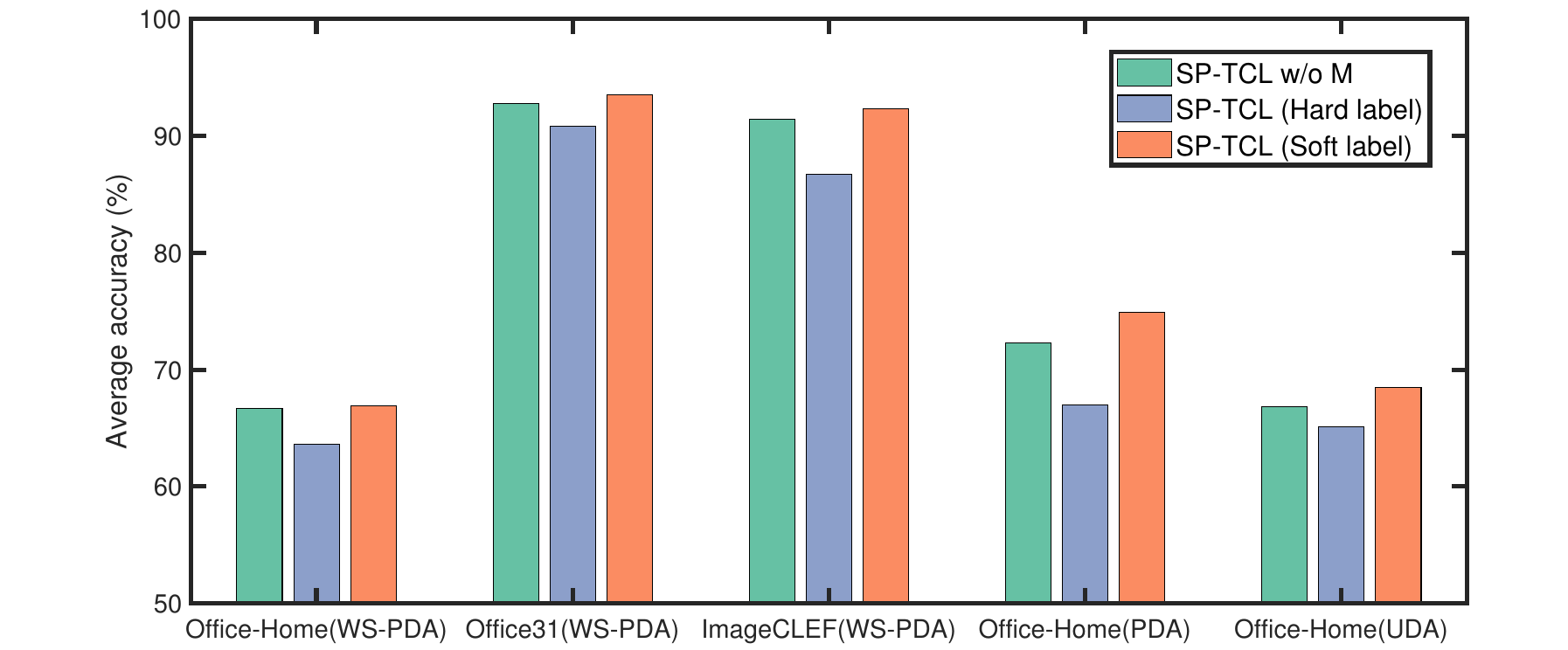}
	\captionsetup{font={footnotesize}}
	\caption{Ablation study of the proposed SP-TCL on four datasets.}
	\label{fig:ablation study}
\end{figure}

\begin{figure*}[t]
	\centering
	\includegraphics[width=6in]{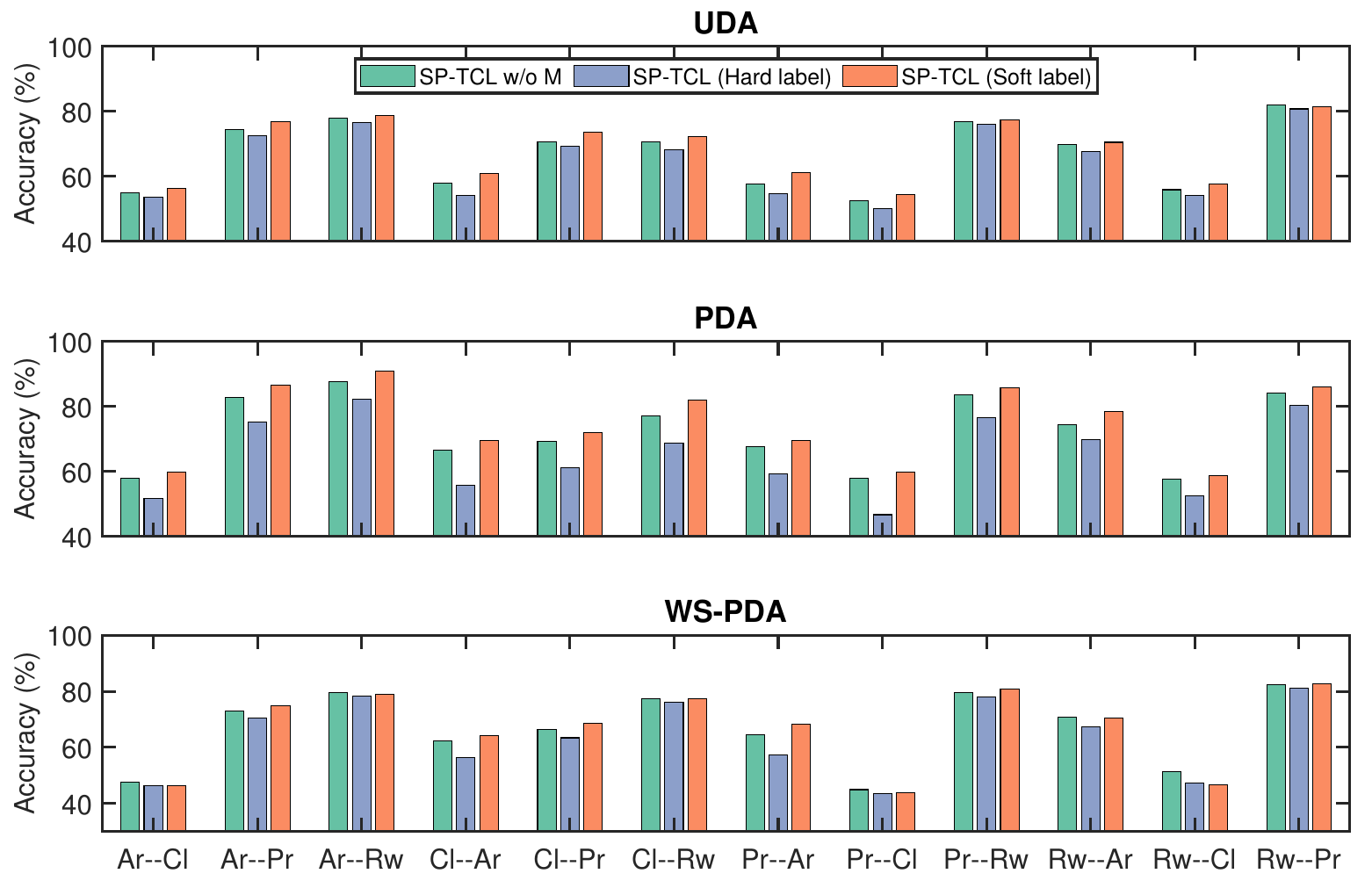}
	\captionsetup{font={footnotesize}}
	\caption{Ablation study of the proposed SP-TCL on the Office-Home dataset under the settings of unsupervised domain adaptation, partial domain adaptation and weakly-supervised partial domain adaptation.}
	\label{fig:ablation study_office_home}
\end{figure*}

\begin{figure*}[t]
	\begin{minipage}[t]{0.33\linewidth}
		\centering
		\includegraphics[width=1.8in]{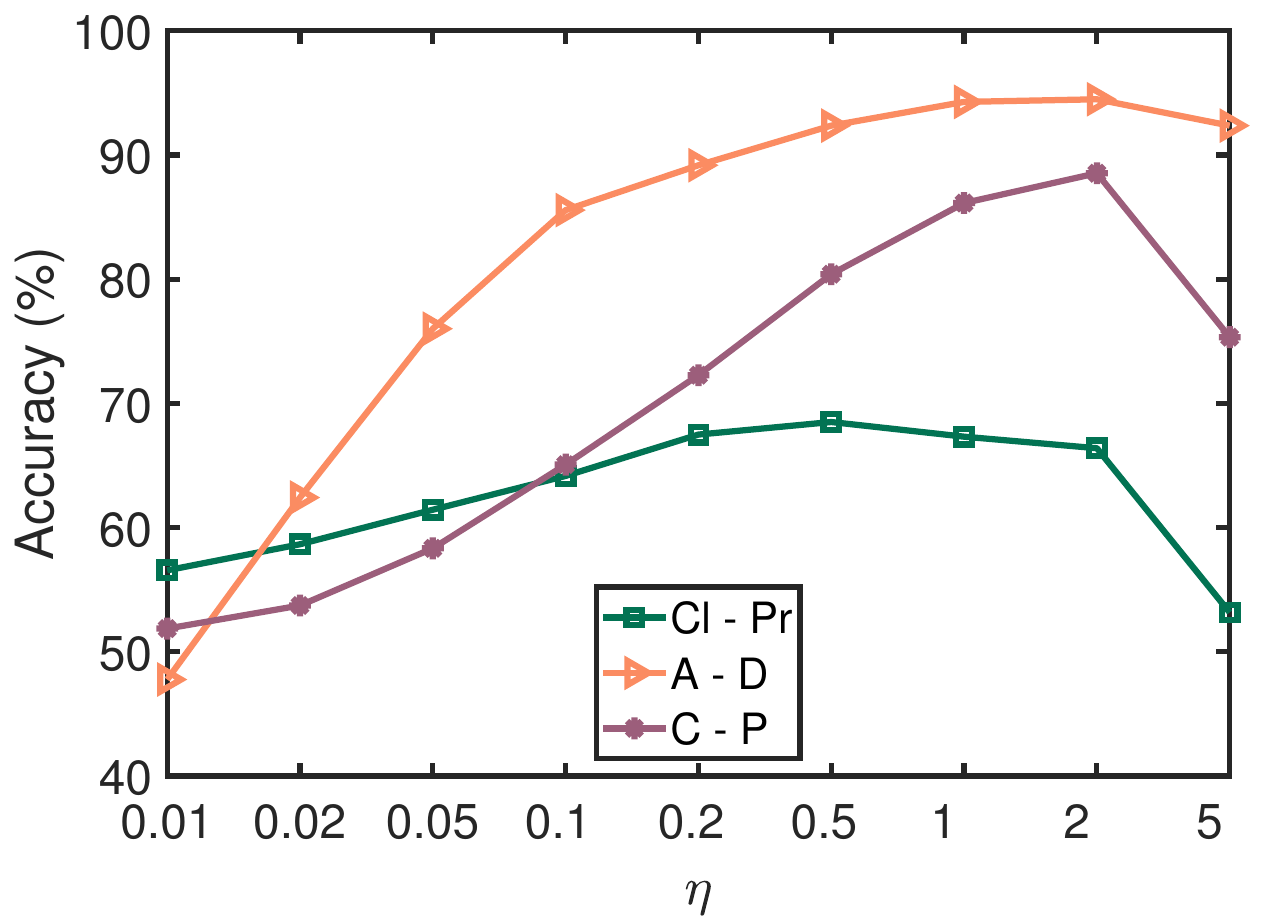}
		\centerline{(a)}
	\end{minipage}
	\begin{minipage}[t]{0.33\linewidth}
		\centering
		\includegraphics[width=1.8in]{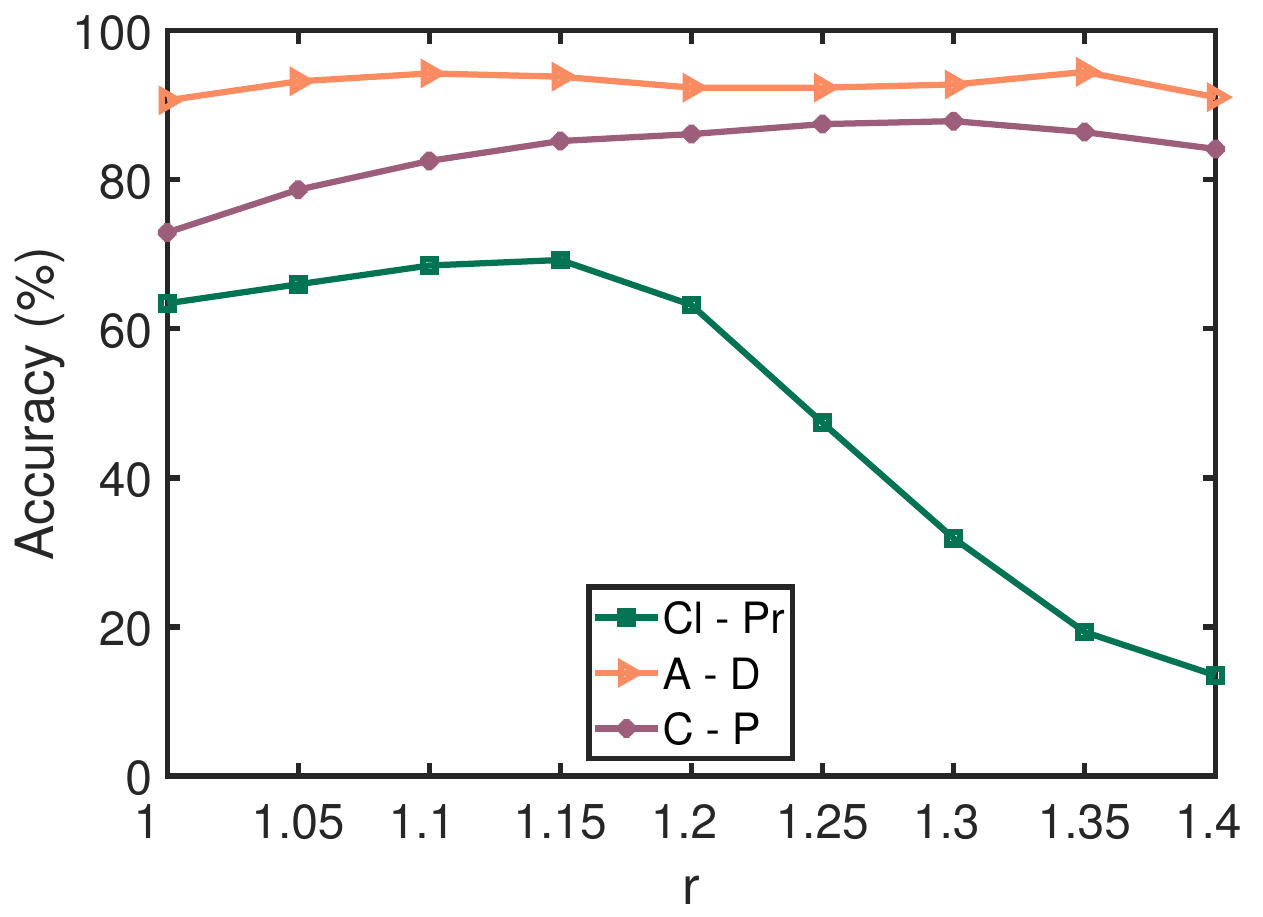}
		\centerline{(b)}
	\end{minipage}
	\begin{minipage}[t]{0.33\linewidth}
		\centering
		\includegraphics[width=1.8in]{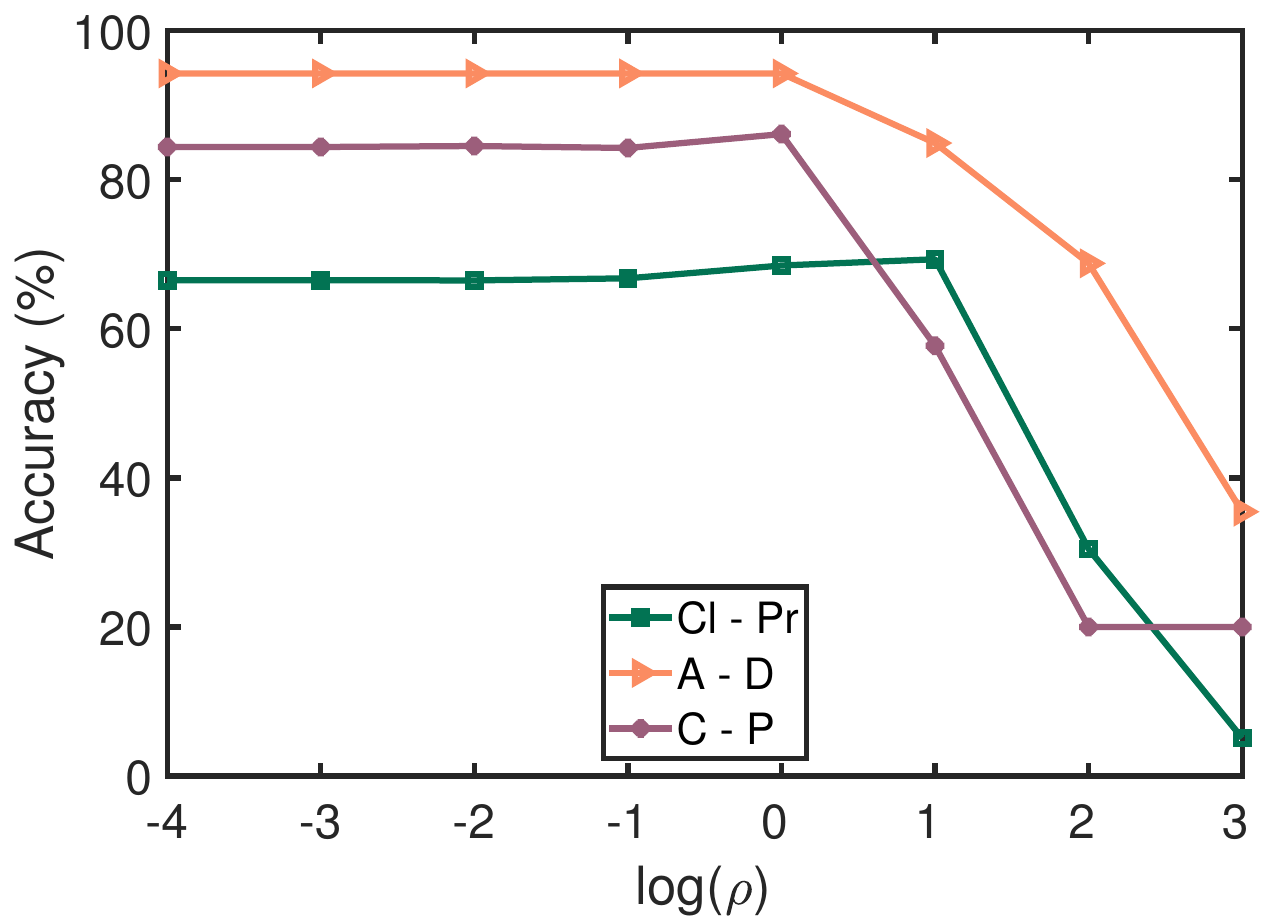}
		\centerline{(c)}
	\end{minipage}
	\captionsetup{font={footnotesize}}
	\caption{Parameter sensitivity on the task Cl$\to$Pr, A$\to$D and C$\to$P. (a) Parameter $\eta$. (b) Parameter $r$. (c) Parameter $\rho$.}
	\label{fig:Parameters study}
\end{figure*}

\subsubsection{Ablation Study}
To dive into the proposed model for in-depth analysis, 
we further perform the ablation study by comparing SP-TCL, termed as SP-TCL (Soft label), with its two variants: 
1) SP-TCL (w/o M) stands for the variant without manifold regularizer;
2) SP-TCL (Hard label) denotes the variant with the square loss $\ell_2$ instead of the prudent loss. 
Experimental results in terms of average accuracy on different datasets under various generalized domain adaptation tasks are illustrated in Fig. \ref{fig:ablation study}.
We also provide the results of each individual task on the Office-Home dataset covering UDA, PDA and WS-PDA tasks in Fig. \ref{fig:ablation study_office_home}. 
From the results in Fig. \ref{fig:ablation study} and \ref{fig:ablation study_office_home}, 
we can see that the performance of SP-TCL (Soft label) in terms of average accuracy is noticeably better than SP-TCL (Hard label).
This is reasonable because the prudent loss function used in our model has potential to adaptively correct noisy source labels and false pseudo target labels,
which can reduce the negative transfer and simultaneously promote the positive transfer.
In addition, SP-TCL (Hard label) only assigns one hard label to each target sample,
which would easily undermine the intrinsic structure of data when there is an overlapping distribution between classes.
These results verify that the prudent loss function is conducive to knowledge discovery.
We also observe that SP-TCL slightly outperforms SP-TCL (w/o M),
proving that manifold regularization can facilitate adaptive target-preferable classifier learning by pushing decision boundaries away from high-density data regions.
Therefore,
manifold regularization can exploit the intrinsic structure of target data to improve  knowledge adaptation across domains.

\subsubsection{Parameter Analysis} \label{exp:para_analysis}
For parameter sensitivity,
there are three tunable parameters in our model:
1) $\eta$ controls the complexity of the decision function $f$;
2) $r$ is used to control the importance of supervised information;
and 3) $\rho$ controls the smoothness of decision function.
We conduct experiments to investigate the sensitivity of these three parameters on three cross-domain tasks: Cl$\to$Pr, A$\to$D and C$\to$P, by varying $\eta\in\{0.01, 0.02, 0.05, 0.1, \cdots, 5\}$, $r\in\{1, 1.05, 1.1, \cdots, 1.4\}$ and $\rho\in\{10^{-4}, 10^{-3},\cdots,10^3\}$.
From Fig. \ref{fig:Parameters study} (a), 
we can see that a small value of $\eta$ would result in poor performance but a proper value of $\eta$ would improve the accuracy. 
It is reasonable that the decision function $f$ will become complicated and then tend to over-fit the noisy source data when $\eta$ is small.
Therefore, $\eta\in[0.5,2]$ is suggested for WS-PDA tasks.
For parameter $r$, theoretically,
when $r$ is too large, $[p^s_{ci}]^r$ and $[p^t_{cj}]^r$ will tend to be zero,
then the supervised information will be lost.
When $r=1$, soft labels will degenerate to hard labels,
which will weaken the adaptability of the model.
From Fig. \ref{fig:Parameters study} (b),
it can be observed that promising results can be achieved when $1<r\le1.2$.
Note that for the cross-domain task Cl$\to$Pr,
the curve decreases quickly when $r\ge1.25$.
That is because Cl$\to$Pr contains more categories (65 classes) and the supervised information will be lost faster,
which would degrade the performance.
Finally, from Fig. \ref{fig:Parameters study} (c), 
we can see that the best results of all cross-domain tasks will be obtained when $\rho=1$,
and the accuracy decreases severely when $\rho>10$.
Therefore, we set $\rho=1$ throughout all our experiments.

\section{Conclusion}\label{sec:conclusion}
In this paper, we present a new approach to weakly-supervised partial domain adaptation,  
a more realistic but much less explored scenario, 
in which we need to transfer a classifier from a large-scale source domain with noises in labels to a small unlabeled target domain. 
Different from existing works that perform knowledge transfer from the perspective of transferable features or instances, 
the proposed method attempts to transfer knowledge from the source domain to the target domain based on the classifier adaptation under the self-paced learning fashion. 
The proposed formulation is general and applicable to various domain adaptation scenarios. 
Convincing results demonstrate the advantage of our approach over the state-of-the-arts on several realistic tasks.


\bibliographystyle{IEEEtran}
\bibliography{pda}


\begin{IEEEbiography}[{\includegraphics[width=1in,height=1.25in,clip,keepaspectratio]{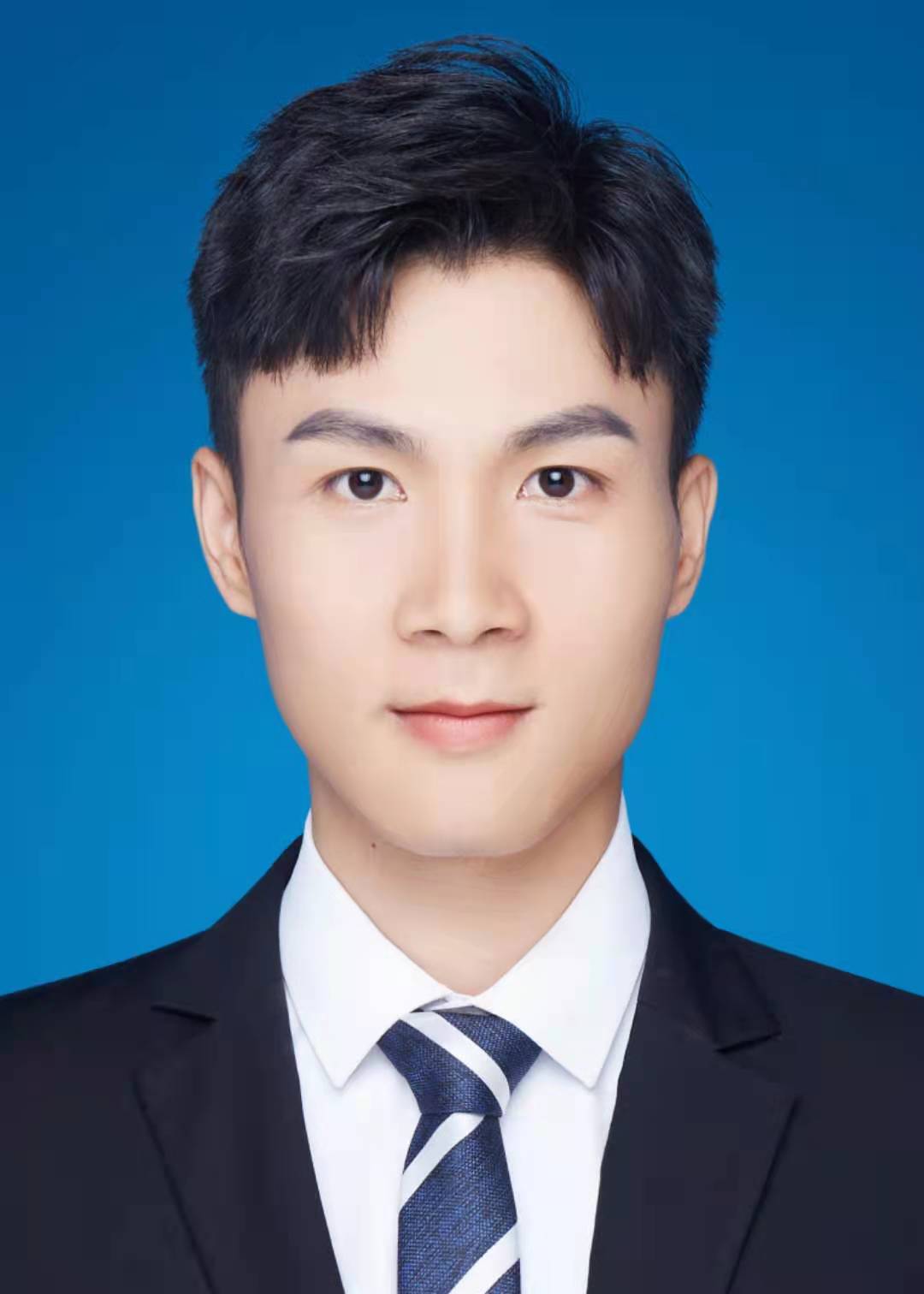}}]{Mengcheng Lan}
	received the B.S. degree from Nanjing University of Aeronautics and Astronautics, Nanjing, China in 2017, and the M.S. degree in computer science and technology from Guangdong University of Technology, Guangzhou, China, in 2020. He is pursuing the Ph.D. degree with the College of Computing and Data Science, Nanyang Technological University, Singapore.
	His current research interests include image processing, semantic segmentation and multimodal large language models. 
\end{IEEEbiography}

\begin{IEEEbiography}[{\includegraphics[width=1in,height=1.25in,clip,keepaspectratio]{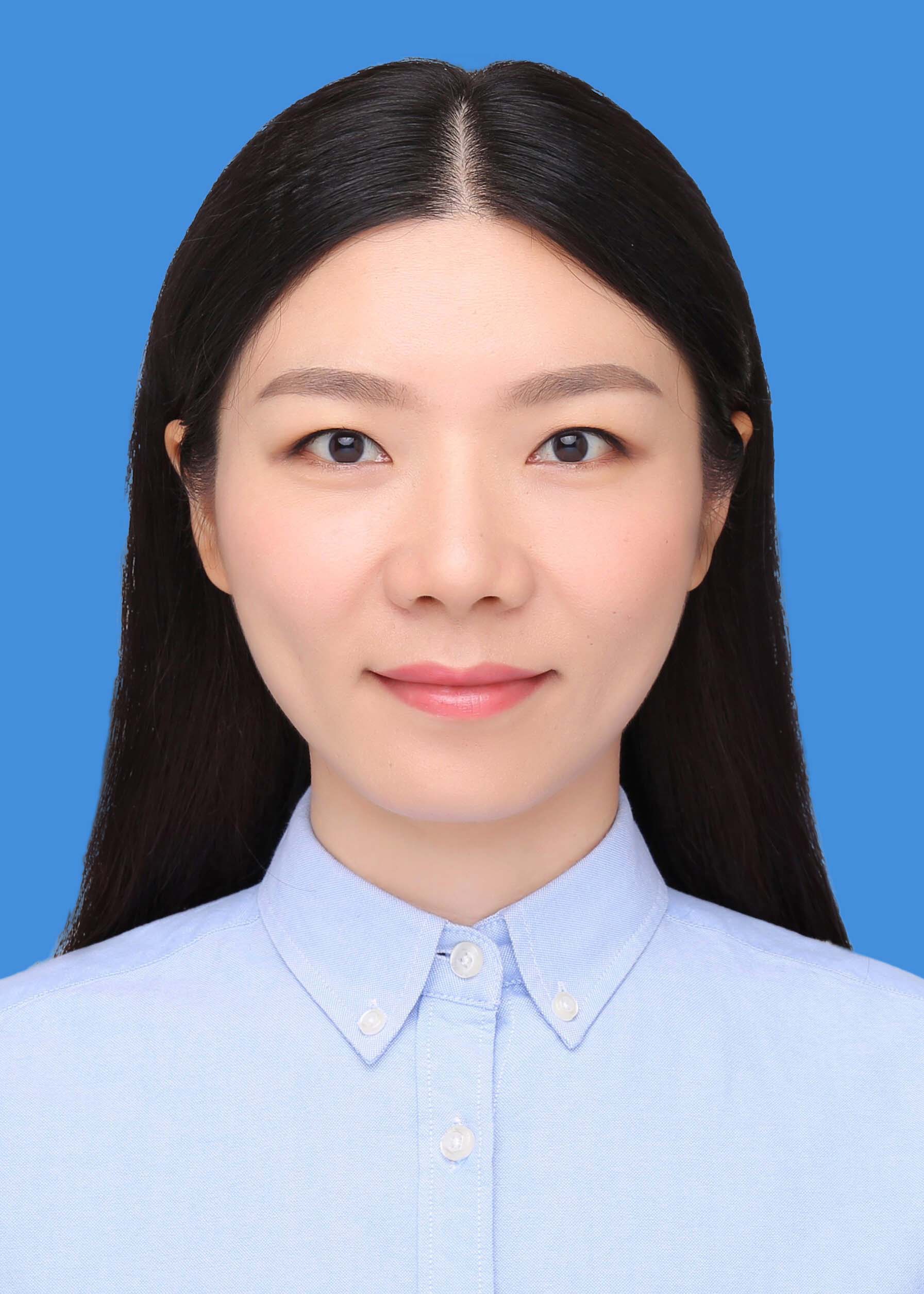}}]{Min Meng}
	(M'17) received the B.S. degree from Zhengzhou University, Zhengzhou, China, in 2006, 
	and the Ph.D degree from Zhejiang University, Zhejiang, China, in 2011. 
	She is currently a Professor with the School of Computer Science and Technology, Guangdong University of Technology, Guangzhou, China. 
	From 2011 to 2013, she was a research fellow with Nanyang Technological University, Singapore. 
	Her current research interests include image processing, multimedia analysis and machine learning. 
\end{IEEEbiography}

\vspace{-1.5in}
\begin{IEEEbiography}[{\includegraphics[width=1in,height=1.25in,clip,keepaspectratio]{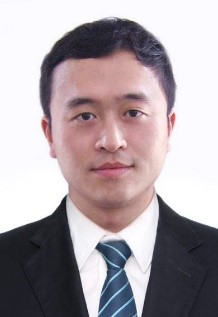}}]{Jun Yu}
	(Senior Member, IEEE) received the B.Eng. and Ph.D. degrees from Zhejiang University, Zhejiang, China.
	He was an Associate Professor with the School of Information Science and Technology, Xiamen University, Xiamen, China. 
	From 2009 to 2011, he was with Nanyang Technological University, Singapore.
	From 2012 to 2013, he was a Visiting Researcher with Microsoft Research Asia (MSRA). 
	He is currently a Professor with the School of Computer Science and Technology, Hangzhou Dianzi University, Hangzhou, China. 
	He has authored or coauthored more than 100 scientific articles. 
	Over the past years, his research interests have included multimedia analysis, machine learning, and image processing.
	In 2017, he received the IEEE SPS Best Paper Award. He has (co-)chaired several special sessions, invited sessions, and workshops. 
	He has served as a program committee member for top conferences including CVPR, ACM MM, AAAI, IJCAI, and has served as associate editors for prestigious journals
	including IEEE Trans. CSVT and Pattern Recognition. 
\end{IEEEbiography}

\vspace{-1.5in}
\begin{IEEEbiography}[{\includegraphics[width=1in,height=1.25in,clip,keepaspectratio]{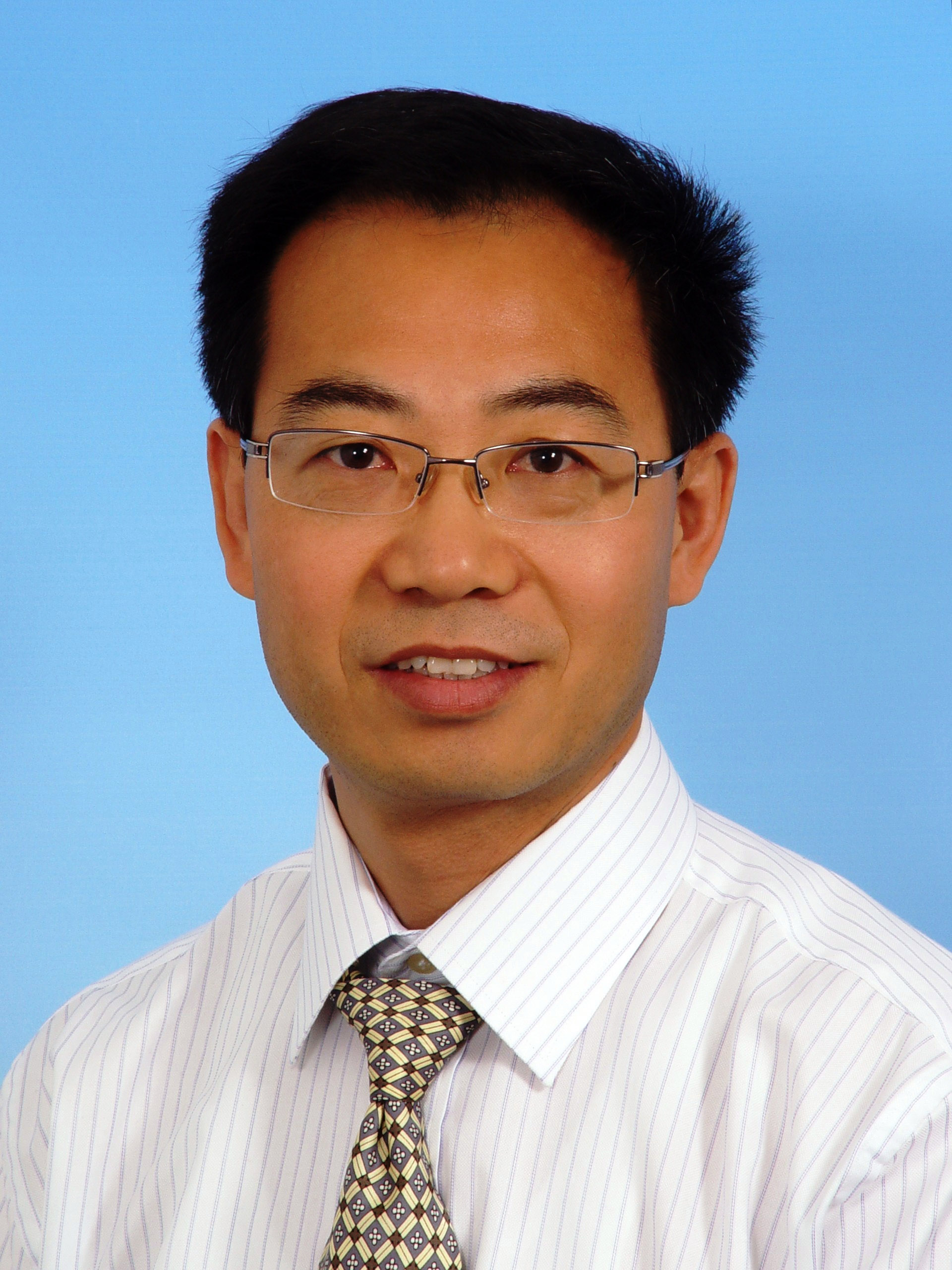}}]{Jigang Wu}
	(M’10) received the B.Sc. degree from Lanzhou University, Lanzhou, China, in 1983, and the Ph.D. degree from the University of Science and Technology of China, Hefei, China, in 2000. He was a Research Fellow with the Center for High Performance Embedded Systems, Nanyang Technological University, Singapore, from 2000 to 2010. He was a Dean and Tianjin Distinguished Professor with the School of Computer Science and Sofware, Tianjin Polytechnic University, Tianjin, China, from 2010 to 2015. He is currently a Distinguished Professor with the School of Computer Science and Technology, Guangdong University of Technology, Guangzhou, China. 
	He has authored over 200 papers in the IEEE TRANSACTIONS ON COMPUTERS, the IEEE TRANSACTIONS ON PARALLEL AND DISTRIBUTED SYSTEMS, the IEEE TRANSACTIONS ON VERY LARGE SCALE INTEGRATION SYSTEMS, the IEEE TRANSACTIONS ON NEURAL NETWORKS AND LEARNING SYSTEMS, the IEEE TRANSACTIONS ON SYSTEMS, MAN, AND CYBERNETICS, the Journal of Parallel and Distributed Computing, Parallel Computing, the Journal of Scientific Achievements, and international conferences. 
	His current research interests include network computing, cloud computing, machine intelligence, and reconfigurable architecture. 
	
	Dr. Wu serves in the China Computer Federation as a Technical Committee Member in the branch committees, high-performance computing, theoretical computer science, and fault-tolerant computing. 
\end{IEEEbiography}

\end{document}